\documentclass[10pt,journal]{IEEEtran}
\usepackage[table]{xcolor}
\usepackage{epsfig}
\usepackage{graphicx}
\usepackage{amsmath}
\usepackage{amssymb}
\usepackage{enumitem}

\usepackage{multirow}
\usepackage{tabularx}
\usepackage{tcolorbox}
\usepackage{dsfont}
\usepackage{url}
\usepackage[tight]{subfigure}
\usepackage{color}
\usepackage{subfigure}
\usepackage{amsthm}
\usepackage[export]{adjustbox}
\usepackage{comment}
\usepackage{diagbox}

\usepackage[utf8]{inputenc} 
\usepackage[T1]{fontenc}    
\usepackage{url}            
\usepackage{booktabs}       
\usepackage{amsfonts}       
\usepackage{nicefrac}       
\usepackage{microtype}      

\usepackage{balance}
\usepackage[rightcaption]{sidecap}

\usepackage{xspace}
\makeatletter
\DeclareRobustCommand\onedot{\futurelet\@let@token\@onedot}
\def\@onedot{\ifx\@let@token.\else.\null\fi\xspace}
\def\eg{\emph{e.g}\onedot} \def\Eg{\emph{E.g}\onedot}
\def\ie{\emph{i.e}\onedot} 
 
\def\etc{\emph{etc}\onedot} 
 
\def\etal{\emph{et al}\onedot}
\makeatother

\definecolor{royalblue}{RGB}{65,105,225} %
\definecolor{lightblue}{RGB}{170,224,250} %
\definecolor{lightgreen}{RGB}{196,223,155}
\definecolor{lightyellow}{RGB}{254,247,153}
\usepackage[pagebackref=true,breaklinks=true,colorlinks,citecolor=royalblue,bookmarks=false]{hyperref}

\usepackage{scalerel}
\usepackage{tikz}
\usetikzlibrary{svg.path}
\definecolor{orcidlogocol}{HTML}{A6CE39}
\tikzset{
  orcidlogo/.pic={
    \fill[orcidlogocol] svg{M256,128c0,70.7-57.3,128-128,128C57.3,256,0,198.7,0,128C0,57.3,57.3,0,128,0C198.7,0,256,57.3,256,128z};
    \fill[white] svg{M86.3,186.2H70.9V79.1h15.4v48.4V186.2z}
                 svg{M108.9,79.1h41.6c39.6,0,57,28.3,57,53.6c0,27.5-21.5,53.6-56.8,53.6h-41.8V79.1z M124.3,172.4h24.5c34.9,0,42.9-26.5,42.9-39.7c0-21.5-13.7-39.7-43.7-39.7h-23.7V172.4z}
                 svg{M88.7,56.8c0,5.5-4.5,10.1-10.1,10.1c-5.6,0-10.1-4.6-10.1-10.1c0-5.6,4.5-10.1,10.1-10.1C84.2,46.7,88.7,51.3,88.7,56.8z};
  }
}
\newcommand\orcidicon[1]{\href{https://orcid.org/#1}{\mbox{\scalerel*{
\begin{tikzpicture}[yscale=-1,transform shape]
\pic{orcidlogo};
\end{tikzpicture}
}{|}}}}

\begin{document}
\title{\emph{FakeLocator}: Robust Localization of GAN-Based Face Manipulations}
\author{Yihao~Huang\,\orcidicon{0000-0002-5784-770X},
        Felix~Juefei-Xu\,\orcidicon{0000-0002-0857-8611},~\IEEEmembership{Member,~IEEE,}
        Qing~Guo\,\orcidicon{0000-0003-0974-9299},~\IEEEmembership{Member,~IEEE,}
        Yang~Liu\,\orcidicon{0000-0001-7300-9215},~\IEEEmembership{Senior~Member,~IEEE,}
        and~Geguang~Pu\,\orcidicon{0000-0001-9750-8334}
\thanks{This work was supported by National Key Research and Development Program (2020AAA0107800), Shanghai Collaborative Innovation Center of Trusted Industry Internet Software, NSFC Project No. 61632005 and NSFC Project. No. 61532019. This work was also supported by the National Research Foundation, Singapore under its the AI Singapore Programme (AISG2-RP-2020-019), the National Research Foundation, Prime Ministers Office, Singapore under its National Cybersecurity R\&D Program (No. NRF2018NCR-NCR005-0001), NRF Investigatorship NRFI06-2020-0001, the National Research Foundation through its National Satellite of Excellence in Trustworthy Software Systems (NSOE-TSS) project under the National Cybersecurity R\&D (NCR) Grant (No.~NRF2018NCR-NSOE003-0001). We gratefully acknowledge the support of NVIDIA AI Tech Center (NVAITC) to our research. 

Yihao~Huang is with East China Normal University, China. Geguang~Pu is with East China Normal University and Shanghai Industrial Control Safety Innovation Technology Co., LTD, China. Felix~Juefei-Xu is with Alibaba Group, USA. Qing~Guo is with Nanyang Technological University, Singapore. Yang~Liu is with Nanyang Technological University, Singapore and Zhejiang Sci-Tech University, China.
Qing Guo and Geguang Pu are the corresponding authors (tsingqguo@ieee.org).}
}

\markboth{IEEE Transactions on Information Forensics and Security}%
{Shell \MakeLowercase{\textit{et al.}}: FakeLocator}

\maketitle

\begin{abstract}
Full face synthesis and partial face manipulation by virtue of the generative adversarial networks (GANs) and its variants have raised wide public concerns. In the multi-media forensics area, detecting and ultimately locating the image forgery has become an imperative task. In this work, we investigate the architecture of existing GAN-based face manipulation methods and observe that the imperfection of upsampling methods therewithin could be served as an important asset for GAN-synthesized fake image detection and forgery localization. Based on this basic observation, we have proposed a novel approach, termed \emph{FakeLocator}, to obtain high localization accuracy, at full resolution, on manipulated facial images. To the best of our knowledge, this is the very first attempt to solve the GAN-based fake localization problem with a gray-scale fakeness map that preserves more information of fake regions. To improve the universality of \emph{FakeLocator} across multifarious facial attributes, we introduce an attention mechanism to guide the training of the model. To improve the universality of \emph{FakeLocator} across different DeepFake methods, we propose partial data augmentation and single sample clustering on the training images. Experimental results on popular FaceForensics++, DFFD datasets and seven different state-of-the-art GAN-based face generation methods have shown the effectiveness of our method. Compared with the baselines, our method performs better on various metrics. Moreover, the proposed method is robust against various real-world facial image degradations such as JPEG compression, low-resolution, noise, and blur.
\end{abstract}


\IEEEpeerreviewmaketitle

\section{Introduction}\label{Introduction}
\begin{figure}
	\centering 
	\includegraphics[width=0.95\columnwidth]{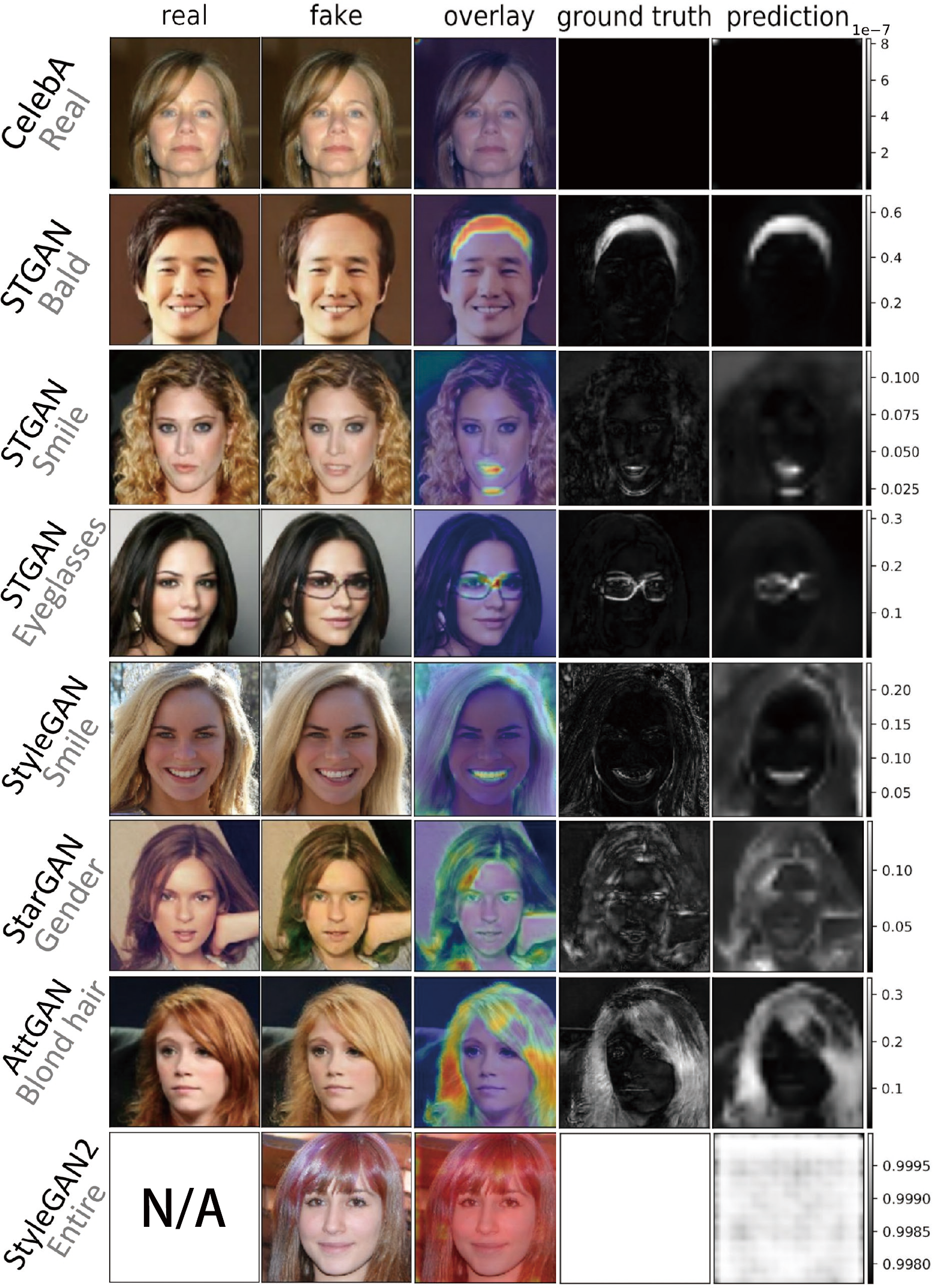}
	\caption{\textbf{Fake region localization results.} These are the results of different GANs and properties. In the left comments, CelebA \cite{liu2018large} is a real image database and others are GAN-based face generation methods. The gray text represents the facial property. 
	\textbf{Fake} image is produced by manipulating corresponding \textbf{real} image through GAN-based face generation method. \textbf{Ground truth} is calculated by \textbf{fake} image and \textbf{real} image. \textbf{Fake} image and \textbf{ground truth} are the input and expected output of our method. \textbf{Overlay} is combined of \textbf{prediction} and \textbf{fake} image. The \textbf{prediction} has colorbar which shows the value range of pixels. For the first row which uses real image of CelebA as input, for unity of the figure, we also regard it as \textbf{fake} image.}
	\label{fig:show}
\end{figure}
Every day we receive newsletters from the media channels such as television, social media, newspaper, \etc. Limited by the presentation of these media, compared with text descriptions, the information with images and videos is prone to be accepted by humans and thus becomes more trustworthy to us. However, with the development of digital manipulation technologies, even videos can be synthesized at a small price. In recent years, a lot of synthetic videos represented by celebrities \cite{rossler2019faceforensics++} were produced by a series of techniques that can produce fake images, audios, and videos, collectively called DeepFakes \cite{DeepFake,juefei2021countering}. DeepFake has been widely used in politics and pornography \cite{fakepolitics,fakepornography}. Fake images can be created easily for that many free tools are available to us. Public concerns about fraud and credibility problems have been raised by such misinformation. Hence, it is urgent to study and investigate effective and robust methods in face forgery detection and forensics, for achieving a safe and responsible multi-media environment.

The facial images synthesized by GAN-based methods are more authentic than other methods. Due to the potential security and privacy issues of synthesized facial images, researchers have a great interest in detecting images generated by GAN-based methods or defending fake generation by tagging the images \cite{wang2021faketagger}. Recently, many studies have worked on classifying images with various methods \cite{yu2019attributing,dang2018deep,mo2018fake,wang2019fakespotter,marra2019gans,wang2020cnn,marra2019incremental,zhou2017two,afchar2018mesonet,guera2018deepfake,matern2019exploiting,agarwal2019protecting,sabir2019recurrent,nguyen2019capsule,sohrawardi2019poster,khodabakhsh2018fake,xuan2019generalization,zhang2019detecting,frank2020leveraging, juefei2021countering}. In these methods, most of them use deep neural networks (DNNs) to deal with the DeepFake detection problem while a few explore other ways. Among the methods using DNNs, the networks used by them can be directly classified into convolutional neural networks (CNNs) and recurrent neural networks (RNNs). In our investigation, \cite{guera2018deepfake} and \cite{sabir2019recurrent} use RNN and take images as the input of the network. Similarly, \cite{dang2018deep,mo2018fake,wang2020cnn,marra2019incremental,zhou2017two,afchar2018mesonet,sohrawardi2019poster,khodabakhsh2018fake,xuan2019generalization,nguyen2019capsule} also take images as input while using CNNs as the backbone networks. In methods that use CNNs, there are three categories based on the input used by them. \cite{yu2019attributing,marra2019gans} consider that DeepFake methods will leave fingerprints in the fake images and they take the fingerprint and images as the input in detection. \cite{zhang2019detecting,frank2020leveraging} take the spectrum of images as the input of CNN. \cite{wang2019fakespotter} uses the activated neuron feature of the given images as the input of DNN to detect DeepFake images. In addition to these DNN-based detection methods, \cite{matern2019exploiting} uses the visual artifacts as the clue. \cite{agarwal2019protecting} models facial expressions and movements that typify an individual’s speaking pattern.

However, none of them considers locating fake regions of the fake images where modifications of facial properties commonly occur. Localization is more significant and valuable in the research field of multimedia forensics.  
In multi-media forensics, a good localization method would better satisfy the following requirements. (1) The localization map is of high resolution with fine-grained fake regions represented (high-resolution). Because it is important to get a good visualization in real forensics scenarios. (2) The method is robust to degradations (robustness), which is important for locators to be deployed in the wild. Because the concept of method robustness is too wide and we can not guarantee to verify all the aspects of method robustness. Thus the ``robustness'' mentioned in the following content mainly represents the degradation robustness. (3) The method is universal enough to tackle unknown facial properties (cross-attribute universality) and unknown GAN methods (cross-method universality). Because the concept of method universality is too wide and we can not guarantee to verify all the aspects of method universality. Thus the ``universality'' mentioned in the following content mainly represents the cross-attribute and cross-method universality.

The generators in GAN-based face generation methods are typical encoder-decoder architecture with an upsampling design in its decoder. The upsampling design is used to magnify the feature maps produced by the encoder to be a colorful image. However, the upsampling design may introduce special features into the synthesized images. According to our investigation, there are only three kinds of upsampling methods. The textures produced by all these upsampling methods contain special features, which we refer to as \textbf{fake texture}. We observe that the \textbf{fake texture} can not only be used for fake detection but also used for fake localization. Thus we propose a universal pipeline that is one of the first attempts to solve the fake localization problem. As an improvement, we also introduce attention mechanism into the architecture to learn \textbf{fake texture} better and generalize our approach to unseen facial properties.

\textbf{FaceForensics++} \cite{rossler2019faceforensics++}, DFFD \cite{dang2020detection}, UADFV \cite{li2018ictu}, and Celeb-DF \cite{li2020celeb} are widely used in DeepFake research. Thus we conduct experiments on them and compare with other state-of-the-art (SOTA) fake detection methods and fake localization methods. 
Furthermore, we build our own dataset with available GANs to generate high-quality forgery images for evaluating the effectiveness and robustness of our proposed method. To the best of our knowledge, only \cite{dang2020detection} and \cite{patchforensics} have been working on the same topic as ours. Therefore, we select their methods as the baselines in our experiment. For \cite{dang2020detection}, in the literature, the authors insert an attention map module into a classifier such as Xception \cite{chollet2017xception} to obtain the location of fake regions. However, the resolution of the attention map is constrained by their design and can only output a tiny map. For example, an image of size 299 $\times$ 299 obtains an attention map of size 19 $\times$ 19. So the attention map can not exactly pinpoint the fake regions. For \cite{patchforensics}, we compare with them on several GAN-based face generation methods. All the localization methods mentioned above do not satisfy high-resolution, universality, and robustness simultaneously. In addition, the fakeness maps of all these localization methods are not suitable for localization task, which will be introduced and improved in our method.

\begin{figure}[tbp]
	\centering 
    \setlength{\belowcaptionskip}{-0.3cm}  
	\includegraphics[width=0.7\columnwidth]{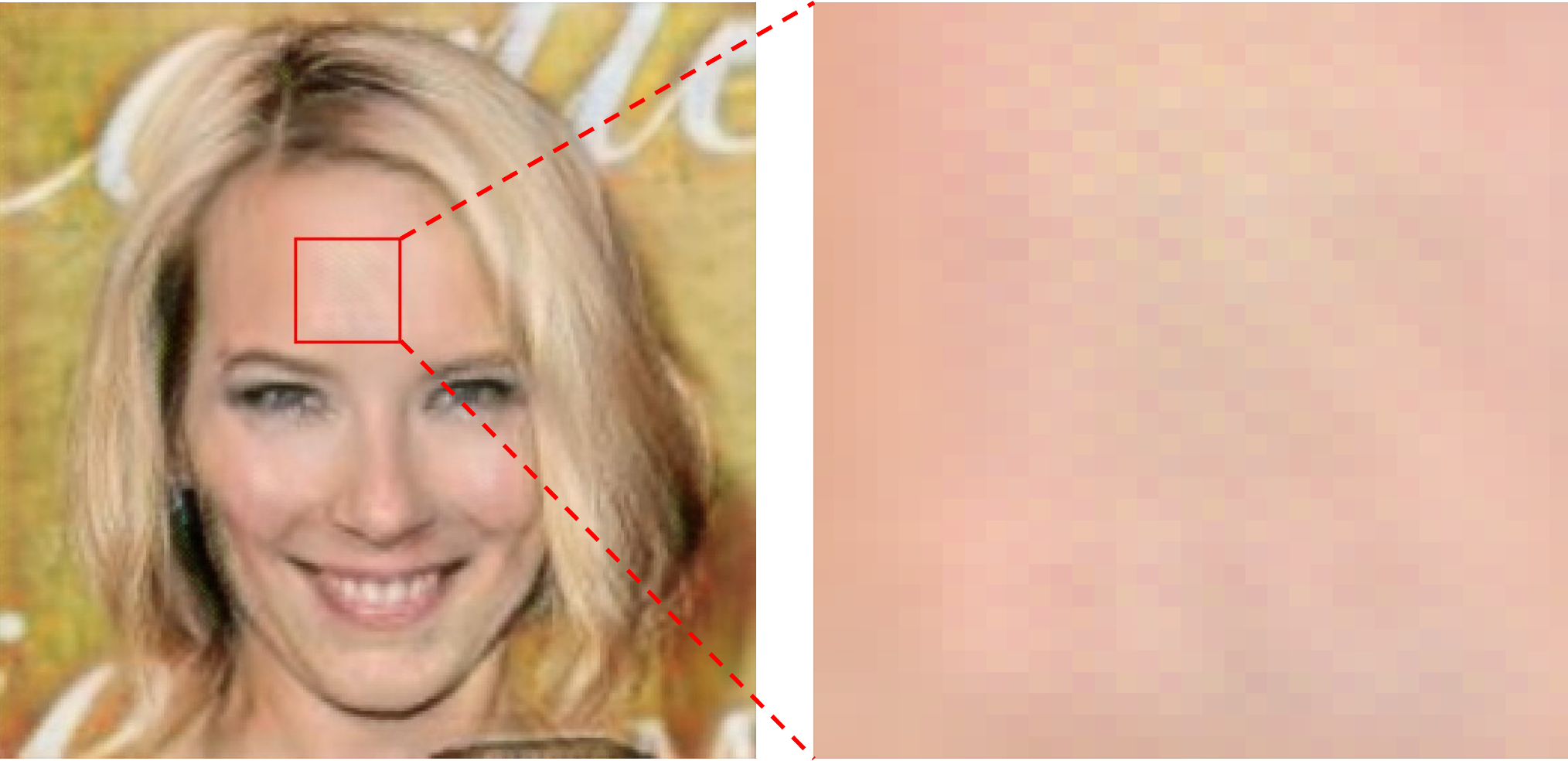}
	\caption{Artifact introduced by GAN-based face generation methods. Here we demonstrate the typical one (\ie, checkerboard pattern). The fake image is produced by StarGAN \cite{choi2018stargan}.}
	\label{fig:checkerboard_pattern}
\end{figure}

The main contributions are summarized as follows.
\begin{itemize}[topsep = 0pt]
    \item We have a new observation that the artifact (shown in Fig.~\ref{fig:checkerboard_pattern}) inducted by GAN-based face generation methods could be used in DeepFake forensics including detection and localization. The pipeline proposed by us can locate the manipulated facial regions effectively at full resolution.
    \item To improve the cross-attribute universality of the model, we introduce the attention mechanism into our framework by using face parsing. To improve the cross-method universality of the model, we propose partial data augmentation and single sample clustering to enhance the training data. The fake textures are captured by the gray-scale fakeness map proposed by us.
    \item Experiments are conducted on two popular DeepFake datasets (\ie, FaceForensics++ and DFFD) and seven SOTA GAN-based face generation methods. Experimental results show that our approach outperforms prior works \cite{dang2020detection} and \cite{patchforensics} in locating fake regions. Furthermore, our method is also robust against various real-world facial image degradations.
\end{itemize}

\section{Related Work}
\subsection{GAN-based Face Generation} 
GAN has drawn attention from both academia and industry since it was first proposed in 2014 \cite{goodfellow2014generative}. The GAN-based face generation methods can be classified into two categories: entire face synthesis and partial face manipulation methods. Here we will introduce seven state-of-the-art GAN-based face generation methods. IcGAN \cite{Perarnau2016}, AttGAN \cite{he2019attgan}, StarGAN \cite{choi2018stargan}, and STGAN \cite{liu2019stgan} are partial face manipulation methods, PGGAN \cite{karras2018progressive} and StyleGAN2 \cite{karras2020analyzing} are entire face synthesis methods. StyleGAN \cite{karras2019style} is not only a partial face manipulation method but also an entire face synthesis method.

IcGAN \cite{Perarnau2016} introduces the encoder that allows the network to reconstruct and modify real face images with arbitrary attributes. PGGAN \cite{karras2018progressive} proposes progressively growing on both the generator and the discriminator to obtain large high-resolution images. StarGAN \cite{choi2018stargan} simply uses a single model to perform image-to-image translations for multiple facial properties. AttGAN \cite{he2019attgan} applies an attribute classification constraint to the generated images to guarantee the correct change of desired attributes. STGAN \cite{liu2019stgan} simultaneously improves attribute manipulation accuracy as well as perception quality on the basis of AttGAN. StyleGAN \cite{karras2019style} proposes a new generator to learn unsupervised separation of high-level attributes and stochastic variation in the generated images. Recently, StyleGAN2 \cite{karras2020analyzing} fixes the imperfection of StyleGAN to improve image quality.

These seven SOTA GAN-based methods typically represent GAN-based entire face synthesis and partial face manipulation methods. Thus we verify the effectiveness of our method on these seven GAN-based methods. In the following sections, \textbf{seven GAN-based face generation methods} are referred to as the GANs introduced here, unless particularly addressed.

\subsection{Manipulated Face Localization}

Although there are a lot of DeepFake detection methods, only several works have been proposed on the manipulated face localization problem. Kritaphat \etal \cite{songsri2019complement} propose an architecture to predict face forensic localization. Huy \etal \cite{nguyen2019multi} use a multi-task learning approach to simultaneously detect manipulated videos and locate the manipulated regions. Li \etal \cite{li2020face} propose face X-ray to locate the fake regions. However, the method fails when the image is entirely synthetic. Furthermore, the datasets used by them are videos that focus on face swap. The fake regions are very large and easy to locate. We mainly focus on locating modified facial properties. Our task is much harder than theirs due to the small fake regions.

Only Joel \etal \cite{dang2020detection} and Chai \etal \cite{patchforensics} have been working on the same topic as us. Dang \etal{} \cite{dang2020detection} present the first and only technique that applies the attention mechanism to address the problem. The attention map is advantageous to be added into arbitrary networks and helps to improve detection accuracy. However, the attention map is too small to point out the fake regions in the fine-grained level. Lucy \etal{} \cite{patchforensics} propose a patch-based classifier with limited receptive fields to visualize fake regions of the images. However, their method puts too much emphasis on local artifacts and ignores global ones.

To sum up, all the localization methods above do not satisfy high-resolution, universality, and robustness simultaneously. Furthermore, the fakeness maps used by these methods may lose tnformation about fake regions.

\begin{figure}[tbp]
	\centering 
    \setlength{\belowcaptionskip}{-0.3cm}  
	\includegraphics[width=0.95\columnwidth]{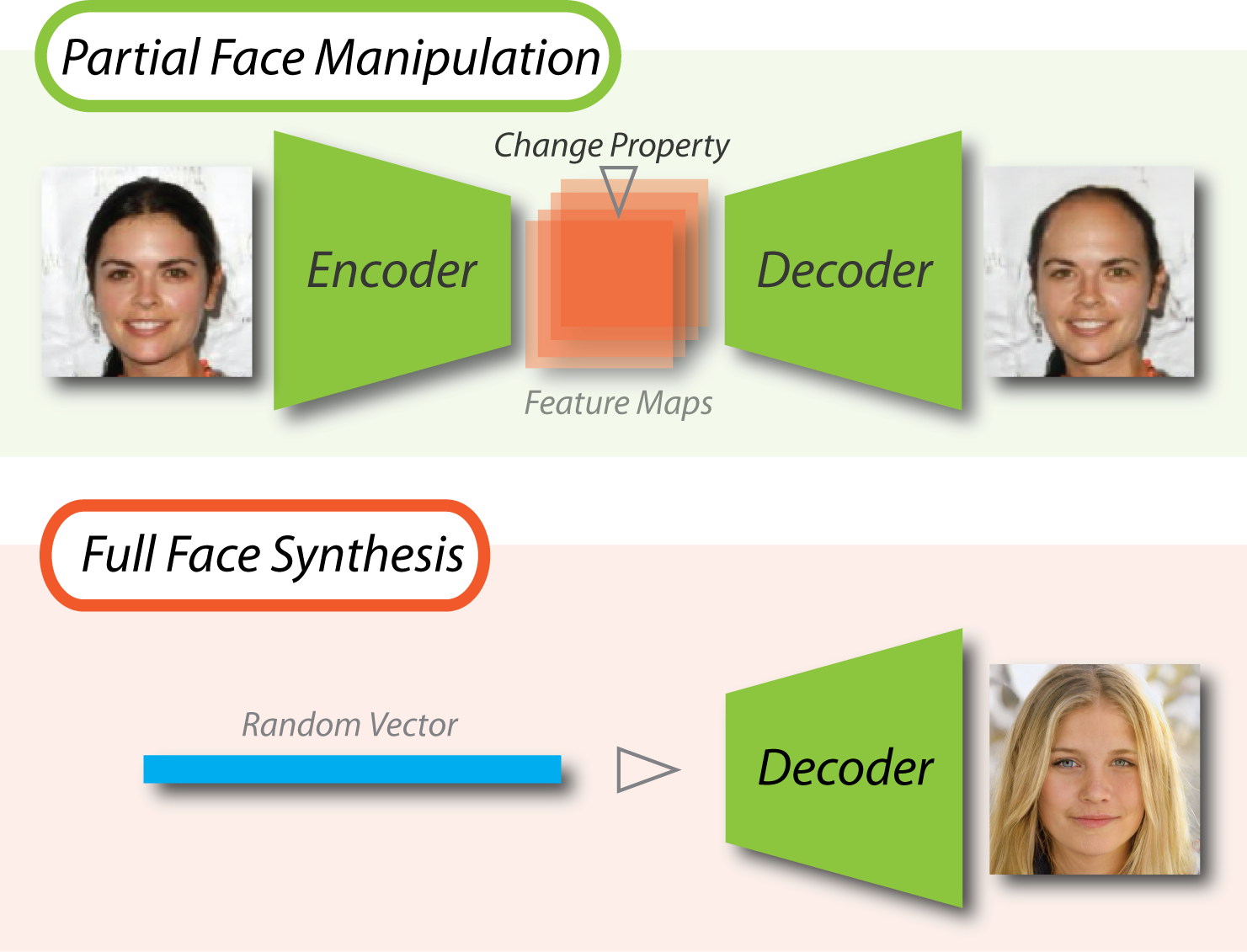}
	\caption{The architecture of GAN-based face generation methods. The top subplot shows how partial face manipulation is carried out. The bottom subplot shows how the entire face synthesis is carried out.}
	\label{fig:GAN_typical_architecture}
\end{figure}
\section{Imperfection of GAN-based methods}
\subsection{General Architecture of GAN-based methods}
There are mainly two ways to generate facial images: entire face synthesis and partial face manipulation. We call these two methods \textbf{face generation methods}. Their typical architecture is based on the encoder-decoder framework shown in Fig.~\ref{fig:GAN_typical_architecture}.

For the partial face manipulation methods, the encoder compresses the real image into small feature maps by convolution and pooling layers. After changing the feature maps with specific facial properties, the modified feature maps are amplified by the upsampling methods in the decoder to be a high-resolution fake image. In the entire face synthesis, the input is a random vector. Similarly, it also needs to go through a decoder to become an entire fake face image. We can find that all the GAN-based face generation methods have the procedure which amplifies the images from low-resolution to high-resolution. In the magnification process, images are inserted with a lot of new pixels calculated by existing pixels. Hence, the output fake face images of these GAN-based methods inevitably contain fake textures that are unlikely obtained from the real world scenario through a camera. 

\subsection{Imperfection of Upsampling}

\begin{figure}[tbp]
	\centering 
    \includegraphics[width=\columnwidth]{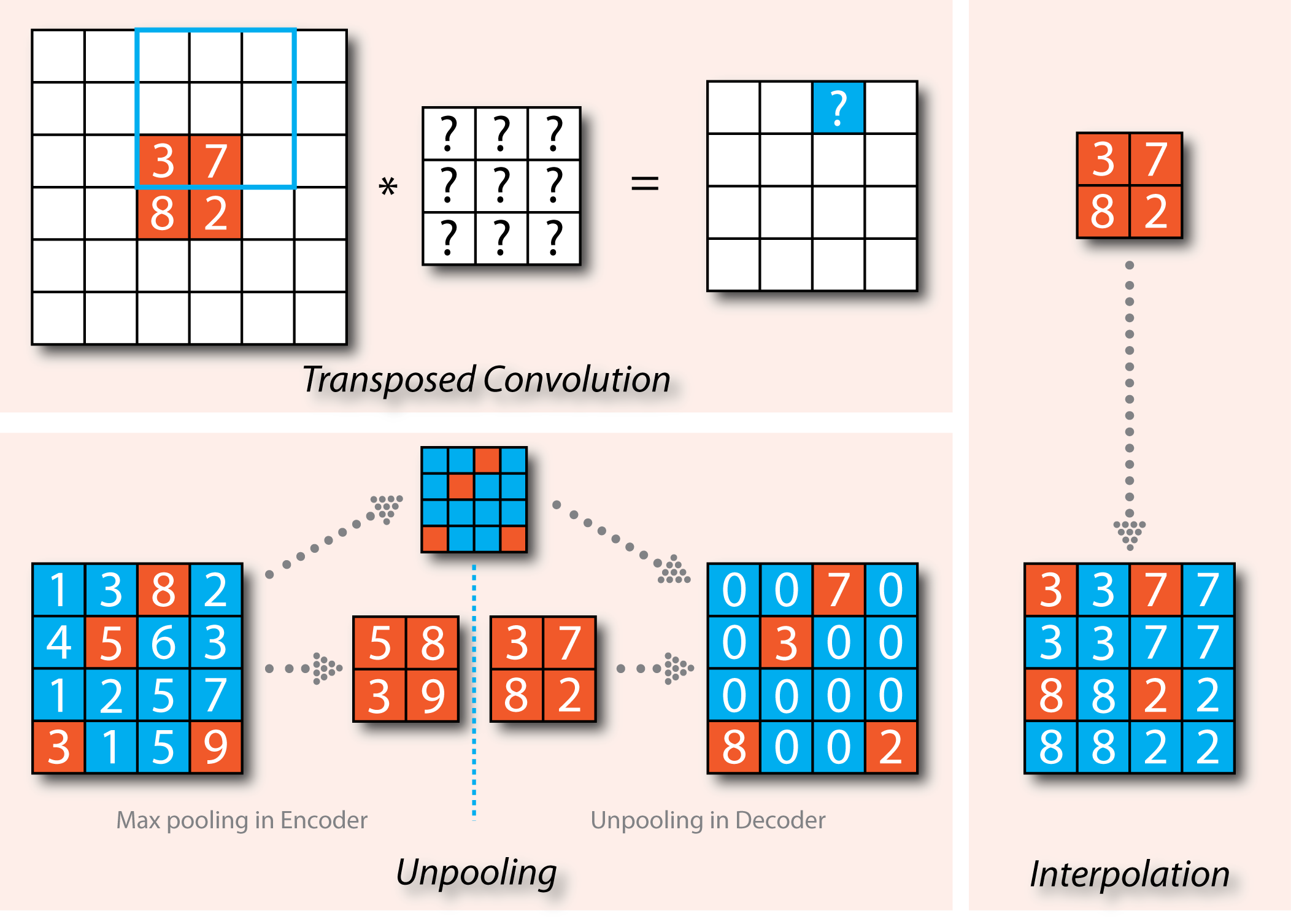}
	\caption{Upsampling methods. Transposed convolution is similar to convolution. It results in the checkerboard texture of the output. In interpolation, the inserted pixels are calculated by the existing pixels. Here we show the nearest neighbor interpolation. Unpooling simply uses zero to fill the inserted pixels, which also produces fake textures.}
	\label{fig:Upsampling methods}
\end{figure}

Upsampling is a technique that can improve image resolution. There are three kinds of upsampling methods: unpooling, transposed convolution, and interpolation, as shown in Fig.~\ref{fig:Upsampling methods}. For GAN-based methods, the most commonly used interpolations are nearest neighbor interpolation, bilinear interpolation, and bicubic interpolation. In Fig.~\ref{fig:Upsampling methods} we just show the nearest neighbor interpolation as an example.

By analyzing the official implementation of \textbf{seven GAN-based face generation methods}, we find that only transposed convolution and interpolation have been used. IcGAN uses interpolation and others use transposed convolution.

Although only two methods are used, all of these three methods have been proved to induce fake texture. Google Brain \cite{odena2016deconvolution} has proved that the transposed convolution results in the checkerboard texture of the output image. \cite{zhang2019detecting} has shown that the transposed convolution and nearest neighbor interpolation have fake texture. Though not mentioned explicitly, the process of their proof also points out that unpooling produces fake textures. For the remaining bilinear and bicubic interpolations, they bring periodicity into the second derivative signal of images \cite{gallagher2005detection}. This means that the interpolated images exhibit fake textures which can be detected by convolution kernel, for instance, Laplace operator.

%
\section{Methodology}

\begin{figure}[tbp]
	\centering
	\includegraphics[width=\linewidth]{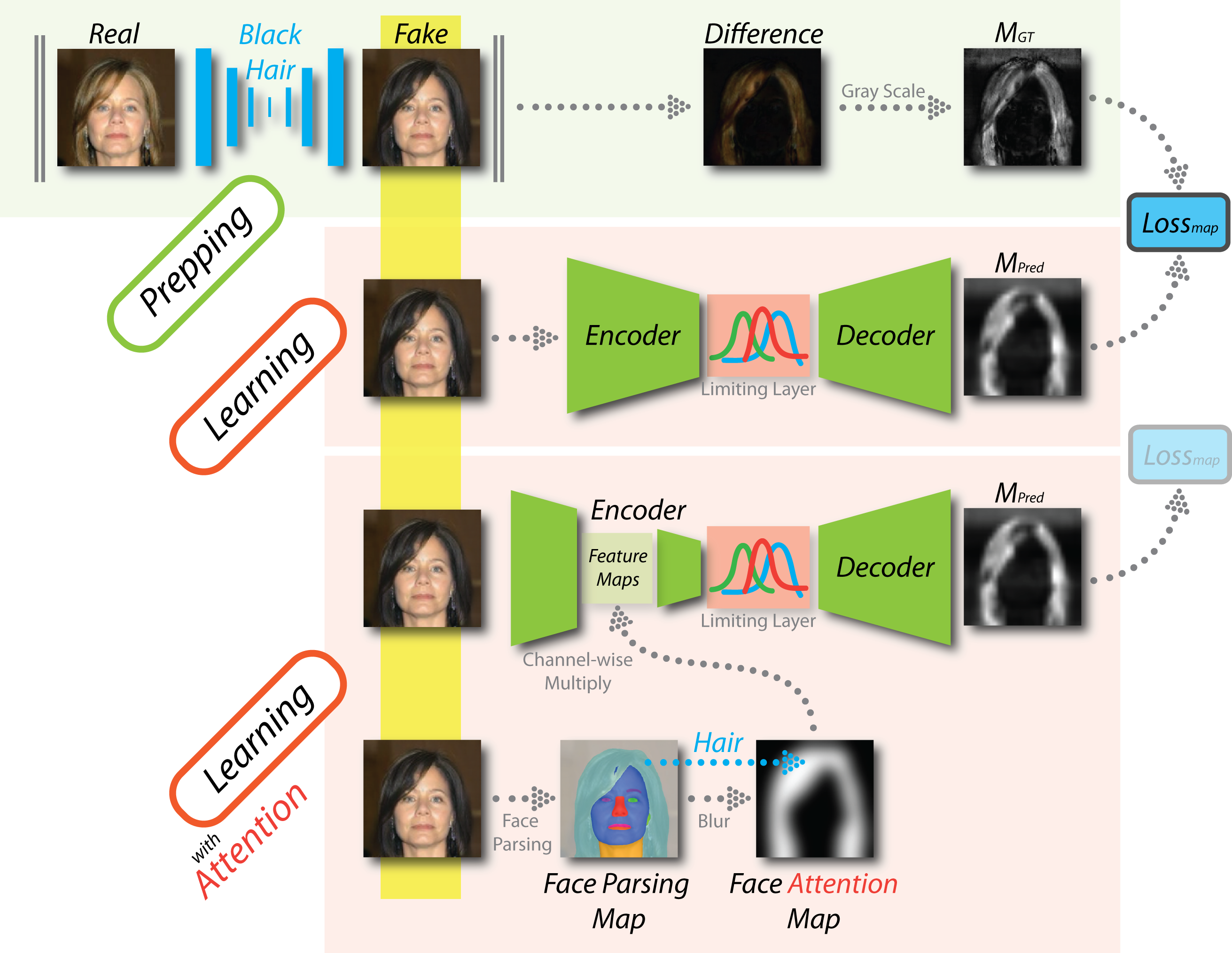}
	\caption{The framework of our proposed \emph{FakeLocator} method. In the prepping procedure, we use pairs of real images and fake images to produce gray-scale fakeness ground truth maps. In the learning procedure, the network is an encoder-decoder architecture. The inputs are real images or fake images while the outputs are gray-scale fakeness prediction maps. We use the gray-scale fakeness prediction maps and the gray-scale fakeness ground truth maps corresponding to the inputs to calculate the loss. To improve the cross-attribute universality of the model, we introduce the attention mechanism into the model by using the face parsing module. We channel-wise multiply the face attention map with the feature maps of the encoder.}
	\label{fig:framework}
\end{figure}


The fake texture produced by upsampling methods is totally different from the real texture in the real image. In this section, we firstly present the framework for face manipulation forensics by leveraging the imperfection of upsampling design in GANs.

Then, we propose the gray-scale fakeness map to visualize the manipulated regions in fake images. The gray-scale fakeness map is more informative than that of other localization methods. To ensure the effectiveness of the gray-scale fakeness map, we adopt a suitable loss function in Sec.~\ref{Loss function}.

Finally, we introduce attention mechanism to generalize our approach in tackling unseen facial properties. We also propose partial data augmentation and single sample clustering to improve the performance on unseen GAN-based fake image generation methods. The pipeline proposed by us produces a full resolution fakeness map. We also verify the robustness and universality of our method in Sec.~\ref{Experiment}.

Please note that since it is almost inevitable for GAN-based face generation methods to use upsampling methods and our method is designed for the imperfection of the upsampling methods, thus our method can be used for localization tasks on most of the GAN-based face generation methods. Furthermore, within a better encoder-decoder network, our method will achieve better performance on fake localization problems.

\subsection{Problem Formulation}\label{problem_formulation}
For fake localization tasks, the main target is to output a fakeness prediction map.
We define the problem as below. The input is a fake image $\mathbf{X^{'}}$ while the output is a fakeness prediction map $\mathbf{M}_{\mathrm{Pred}}$. The solution is a function $\mathcal{F}(\cdot)$.

\begin{equation}
\mathbf{M}_{\mathrm{Pred}}= \mathcal{F}(\mathbf{X^{'}}).
\end{equation}

Furthermore, we think that the method should satisfy three properties. First, the output fakeness prediction maps should have the same solution as the fake counterparts. Second, as we mainly locate the fake regions of the face, the cross-attribute universality of the method on different facial properties is necessary. That means, if we use images that change hair color to train a model, the model should have the ability to locate the fake regions of other facial properties manipulated. Meanwhile, as different GAN methods generate various fake textures, the cross-method universality of the method is necessary as the supplementary to cross-attribute universality. Third, the method should be robust to many degradations that may be used by attackers and evaders.

To output a full resolution fakeness prediction map, the method should provide fake images and their ground truth maps for training. The generation method of the fakeness prediction map should fully reflect the location and strength of the fake region. This requires a good design of fakeness maps. In Sec.~\ref{Implementation}, we introduce the implementation of our design and the drawbacks of other methods. 

The cross-attribute universality of different facial properties is a difficult problem. Existing GAN-based face generation methods inevitably modify the pixels in other regions that are outside the regions of modified facial properties. Therefore, they use some methods (\eg, skip connection) to repair the fake images, which makes the region outside modified properties not totally fake. Only the regions of modified properties are full of fake textures. Thus the textures learned by the model in the training procedure lose the universality. To solve this problem, we propose a face-aware attentional encoder-decoder network to improve the cross-attribute universality of the model. 

The cross-method universality of various GAN methods is also a difficult problem. To improve the performance of the model across different GAN methods, we propose partial data augmentation and single sample clustering in Sec.~\ref{partial data augmentation}. In particular, we only add data augmentation to the real images of the training dataset while keeping fake images unchanged. The effectiveness of this method is shown in Table~\ref{tab:ablation_study_general_test} and Table~\ref{tab:augmentation_general_test} of Sec.~\ref{Experiment}.

In the real world, images may be degraded by various operations such as compression, low-resolution, \etc. We fully verify the robustness of the model to see whether the proposed method can survive these degradations in Sec.~\ref{Experiment} .
\begin{figure}[tbp]
	\centering 
	\includegraphics[width=0.8\columnwidth]{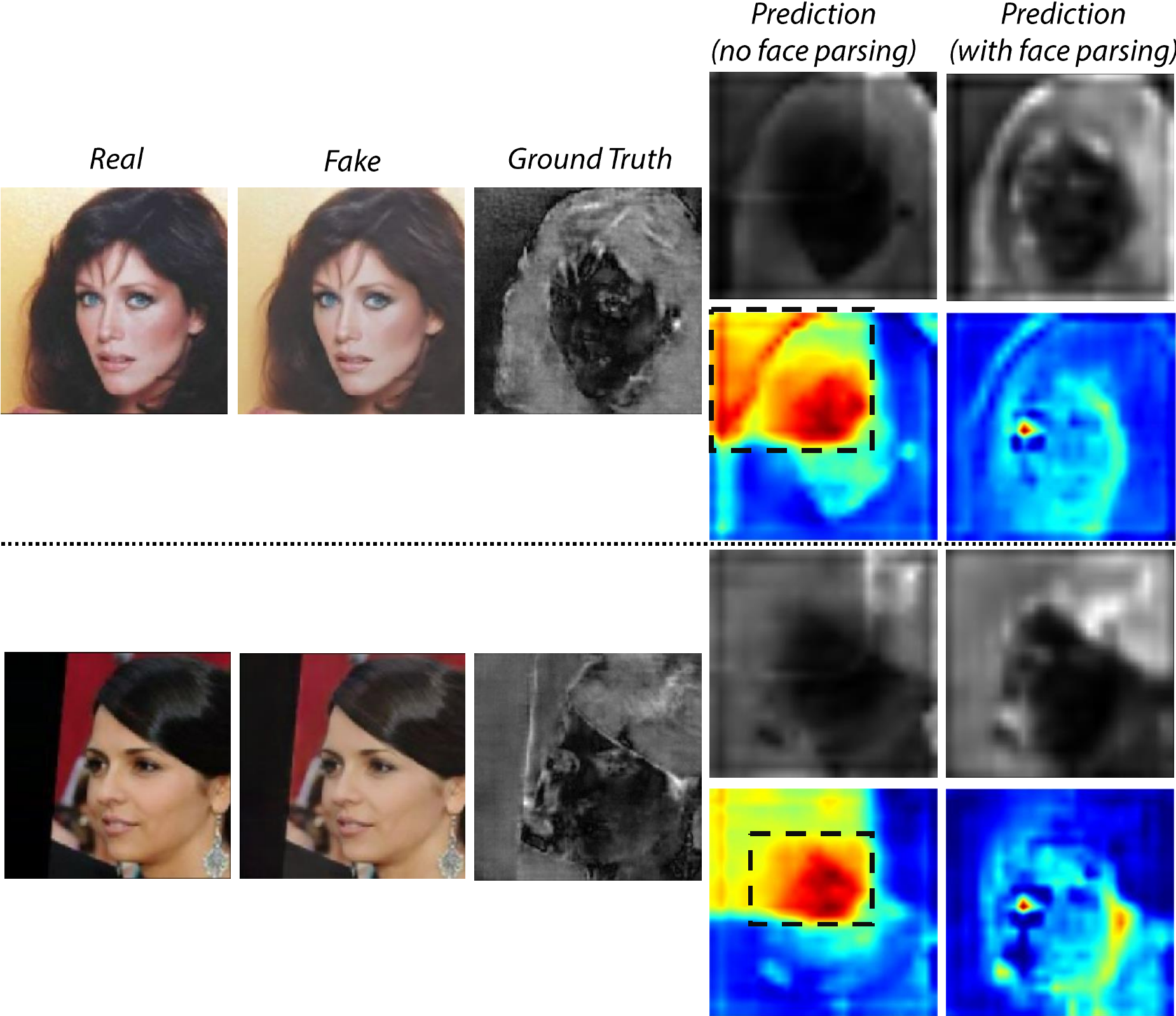}
	\caption{The images from left to right: real image, fake image, ground fakeness map, (top: prediction fakeness map, bottom: attention map) of the model without face parsing module trained by the property \emph{Blond hair}, (top: prediction fakeness map, bottom: attention map) of the model with face parsing module trained by the property \emph{Blond hair}.}
	\label{fig:face_parsing_feature_compare}
\end{figure}

\subsection{General Encoder-Decoder Network for Fake Localization}\label{sec:general_encoder_decoder_network}
Fig.~\ref{fig:framework} shows the framework of our methodology. The backbone network is an encoder-decoder architecture. In our method, any encoder-decoder network can be used as the backbone network. The input is a fake image $\mathbf{X^{'}}$ while the output is a fakeness prediction map $\mathbf{M}_{\mathrm{Pred}}$. 
\begin{equation}
\mathbf{M}_{\mathrm{Pred}}= \mathcal{F}_{\mathrm{dec}}(\mathcal{F}_{\mathrm{enc}}(\mathbf{X^{'}})).
\label{prediction_map}
\end{equation}
$\mathcal{F}_{\mathrm{dec}}$ and $ \mathcal{F}_{\mathrm{enc}}$ are the decoder and encoder of \emph{FakeLocator} respectively. $\mathbf{M}_{\mathrm{GT}}$ represents the fakeness ground truth map. In our method, the pixel values in the fakeness prediction map are real numbers. So we should fine-tune the network and limit the pixel values in the fakeness prediction map. We add a limiting layer between the encoder and the decoder so that the pixel values could be limited to the range within [0,1]. The subsequent decoder can further process these intermediate features without worrying about correcting different scales of the features. This renders the learning of the decoder much more efficient. This limiting layer can also be added after the decoder. The loss $ \mathcal{L}_{\mathrm{map}}$ is simply calculated by the following formula and we will detail the $\mathcal{L}$ in Sec.~\ref{Loss function}.
\begin{equation}
    \setlength\abovedisplayskip{3pt}
    \setlength\belowdisplayskip{4pt}
    \mathcal{L}_{\mathrm{map}}  = \mathcal{L}(\mathbf{M}_{\mathrm{Pred}},\mathbf{M}_{\mathrm{GT}}).
\end{equation}

We will introduce the network architectures of $ \mathcal{F}_{\mathrm{dec}}$ and $ \mathcal{F}_{\mathrm{enc}}$ in Sec.~\ref{Experimental_Setup}. 
Furthermore, to allow fake classification, we add a binary classifier as a new branch after the encoder and formulate this process with the input image $\mathbf{X}^{'}$ as
\begin{equation}
y_{\mathrm{pred}}= \mathcal{C}(\mathcal{F}_{\mathrm{enc}}(\mathbf{X^{'}})),
\label{cls_pred}
\end{equation}
%
where $\mathcal{C}$ denotes the classification branch and $y_{\mathrm{pred}}\in\{\text{real},\text{fake}\}$ is the predicted category of the $\mathbf{X^{'}}$. 
With this classification module, we can calculate the fake classification accuracy that is an important metric to validate the discriminative power of the encoder and makes our method comparable to state-of-the-art DeepFake detection methods.
We will detail the architecture of the classification branch in Sec.~\ref{Experimental_Setup}.

Although the above framework has the ability to locate fake regions, the cross-attribute universality still needs improvement.
As shown in Fig.~\ref{fig:face_parsing_feature_compare}, we use two groups of images to demonstrate the issue in cross-attribute detection. In both groups, the fake images in Fig.~\ref{fig:face_parsing_feature_compare} are generated by manipulating real images with the property \emph{Black hair}. The model used to locate fake regions is trained by the property \emph{Blond hair}. In the column ``prediction (no face parsing)'', the top image is the prediction result of the previous model while the bottom image is the attention of the model on the fake image. The prediction result is not good enough. We can find that the model lacks the cross-attribute universality for that it puts too much attention on the face area (\ie, regions enclosed by the dotted line).

\begin{figure}[tbp]
	\centering 
	\includegraphics[width=\columnwidth]{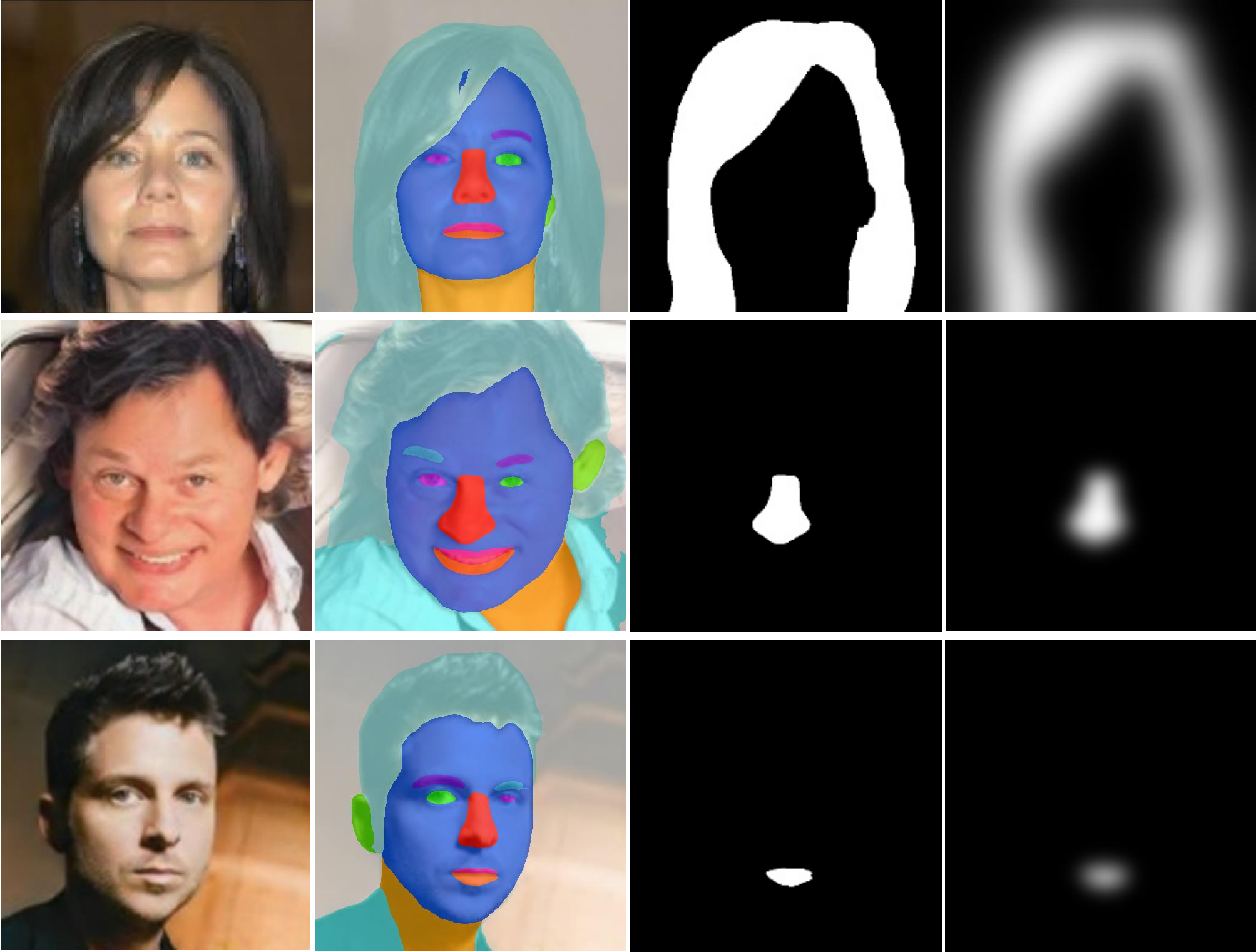}
	\caption{From left to right: fake image, face parsing map, region map of the modified property, face attention map. The emphasized properties from top to bottom are \emph{Hair}, \emph{Nose}, and \emph{Mouth}.}
	\label{fig:face_parsing_map}
\end{figure}

\subsection{Face-aware Attentional Encoder-Decoder Network}\label{Face parsing}
The previous method obtains a good result in detecting and locating the fake images of the specific facial properties. That is, the model trained with images that have facial property $p$ changed has a good performance on the corresponding fake images while performing poorly on detecting and localizing images with facial property $p'$ ($p' \neq p$) changed.

Since the decoder of each trained GAN-based fake image generation model is fixed, the fake texture generated by the same decoder should be similar. Under this situation, we think that the problem is caused by the following reason. Although the GAN-based face generation methods can change facial properties very well, they still inevitably modify the pixels in other regions of the image. Therefore, they use some methods (\eg, skip connection) to repair the fake images by referring to the real ones, which destroys the fake texture of the region outside modified properties and makes the region not totally fake. Only the regions of modified properties are full of fake textures. Thus, we should provide additional information to make the network pay more attention to the fake texture in the regions of modified properties. Thus we introduce attention mechanism into our method by using a face parsing module to learn the fake textures better. The experiment in Table~\ref{tab:compare_face_parsing} demonstrates the effectiveness of the face parsing module.

As shown in Fig.~\ref{fig:framework}, we use face parsing to mark the area corresponding to the modified property and insert the face attention map into the encoder. The attention mechanism urges the model to put more emphasis on the regions of modified property, which improves the cross-attribute universality of the model. In the training step, the face attention map is calculated by the face parsing module ($\mathrm{FPM}$). In the testing step, the face attention map is a white map that does not provide any region information and reserves the original information of feature maps in the encoder. 
If we set the modified facial property as $p$, then the face parsing map of fake image $\mathbf{X^{'}}$ is $\mathrm{FPM}(\mathbf{X^{'}},p)$, where function $\mathrm{FPM}(\cdot)$ represent the face parsing operation of the facial image. The face attention map can be written as $\mathrm{Blur}(\mathrm{FPM}(\mathbf{X^{'}},p))$, where $\mathrm{Blur}(\cdot)$ means the blur operation. The formula of the encoder which takes face parsing module into account shall be rewritten as ($\mathcal{F}_{\mathrm{enc}}(\mathbf{X^{'}},\mathrm{Blur}(\mathrm{FPM}(\mathbf{X^{'}},p)))$). Eq.~\eqref{prediction_map} shall be rewritten as 
\begin{equation}
\mathbf{M}_{\mathrm{Pred}}= \mathcal{F}_{\mathrm{dec}}(\mathcal{F}_{\mathrm{enc}}(\mathbf{X^{'}},\mathrm{Blur}(\mathrm{FPM}(\mathbf{X^{'}},p))))).
\end{equation}

As shown in Fig.~\ref{fig:face_parsing_map}, we demonstrate the face attention map corresponding to the modified properties. The images in turn are the fake images, the face parsing maps, the region maps of the modified properties, and our proposed face attention maps. To produce the face attention map, there are three steps. First, use a face parsing method to generate a face parsing map with the input image. Second, choose the region of the modified property. Third, use the blur method to expand the white region. The reason why we add the third step is that the region outside modified properties also has reference significance. However, their weight should not be higher than the regions of modified properties. The blur method is suitable for this transformation.

As shown in the last column of Fig.~\ref{fig:face_parsing_feature_compare}, the model with face parsing module trained by the property \emph{Blond hair} shows better results on locating fake regions. We can find that the model does not put too much emphasis on the face area. Instead, it distributes enough attention to other areas, which makes it easier to locate the fake textures in the hair area.

\subsection{Partial Data Augmentation}\label{partial data augmentation}
To evaluate the cross-method performance of our method, we compare on three DeepFake types: entire face synthesis, attribute manipulation, identity swap. As shown in Table~\ref{tab:augmentation_general_test_diff_DeepFake_category}, the models in the first column are trained by the corresponding datasets (\ie, FaceForensics++, StarGAN, StyleGAN, PGGAN). The datasets in the first row are the test datasets. In each cell, the left value is the accuracy result. We can find that the cross-method accuracy between attribute manipulation and entire face synthesis are high while the cross-method accuracy between them and identity swap is pretty low.

\begin{table*}[tbp]
\centering
\caption{The accuracy between three different DeepFake types. The first column shows the models while the second row shows the test datasets. In each cell, the left value is the accuracy of the model without data augmentation and the second value is the accuracy of the model with partial data augmentation. The third value (if exists) is the accuracy of the model with partial data augmentation and single sample clustering.}
\label{tab:augmentation_general_test_diff_DeepFake_category}
\begin{tabular}{l|c|c|c|c}
\toprule 
 \multirow{2}{*}{\diagbox{Model}{Dataset}}& \multicolumn{1}{c|}{identity swap} & \multicolumn{1}{c|}{attribute manipulation} & \multicolumn{2}{c}{entire face synthesis}\tabularnewline
 & FaceForensics++ & StarGAN & StyleGAN & PGGAN\tabularnewline
\midrule
 FaceForensics++& - & 0.490/0.988 (0.999) & 0.363/0.500 (0.854) & 0.369/0.500 (0.975)\tabularnewline
 StarGAN& 0.425/0.449 (0.817) & - & 1.000/0.999 & 1.000/0.999\tabularnewline
 StyleGAN& 0.452/0.452 (0.817) & 0.989/0.993 & - & 1.000/0.999\tabularnewline
PGGAN& 0.448/0.451 (0.676) & 0.999/0.990 & 1.000/0.999 & -\tabularnewline
\bottomrule 
\end{tabular}  
\end{table*}

Intuitively, we can take data augmentation into consideration. However, simply using data augmentation is not effective. We perform data augmentation on FaceForensics++ model and test it on StarGAN, StyleGAN, and PGGAN. The accuracy is 0.454, 0.449 and 0.411, respectively, which is only a little higher than without augmentation.

Thus we propose partial data augmentation to further improve the cross-method ability of the model. We think that using data augmentation on both real and fake images may confuse their distribution. Thus we only do data augmentation on real images to tell the model the maximum possible distribution of real images, which instructs it to treat any image that is not real as fake. As shown in Table~\ref{tab:augmentation_general_test_diff_DeepFake_category}, the second value of each cell is the accuracy result of partial data augmentation. We can find that the accuracy is higher than without data augmentation and performing data augmentation on both real and fake images. However, the accuracy result between identity swap and the other two DeepFake types is still not high enough.

\begin{figure}[tbp]
	\centering 
	\includegraphics[width=\columnwidth]{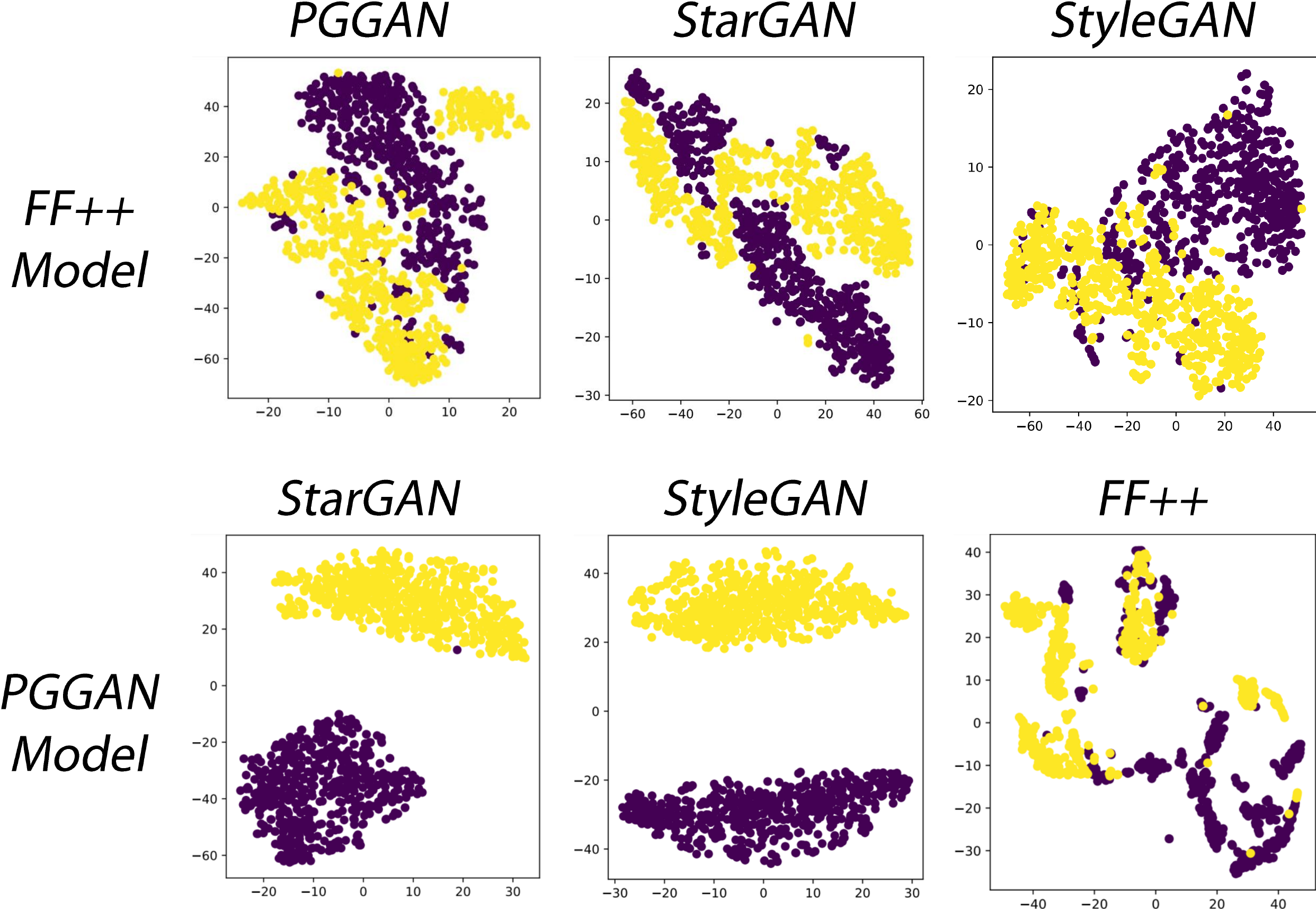}
	\caption{The t-SNE visualization of features before classifier. The models are on the left and the test datasets are above the right three subfigures.}
	\label{fig:t-sne_embedding_with_no_aug}
\end{figure}

We further research the features before classifier (\ie, the feature output of the encoder). To be specific, we use t-SNE \cite{van2008visualizing} to reduce the dimension of features and visualize them. As shown in the first row of Fig.~\ref{fig:t-sne_embedding_with_no_aug}, we use t-SNE to demonstrate the feature of the FaceForensics++ model on PGGAN, StarGAN, StyleGAN. We can find that the features of real images are entangled with fake images and it is hard to separate them. In the second row of Fig.~\ref{fig:t-sne_embedding_with_no_aug}, we use t-SNE to demonstrate the feature of the PGGAN model on StarGAN, StyleGAN, FaceForensics++. We can find that the FaceForensics++ features of real images are entangled with fake images while the real and fake images features of StarGAN and StyleGAN are clearly distinct. The visualization is consistent with the experiment result that the cross-method accuracy between attribute manipulation type and entire face synthesis type are high while the cross-method accuracy between them and identity swap is pretty low.

\begin{figure}[tbp]
	\centering 
	\includegraphics[width=\columnwidth]{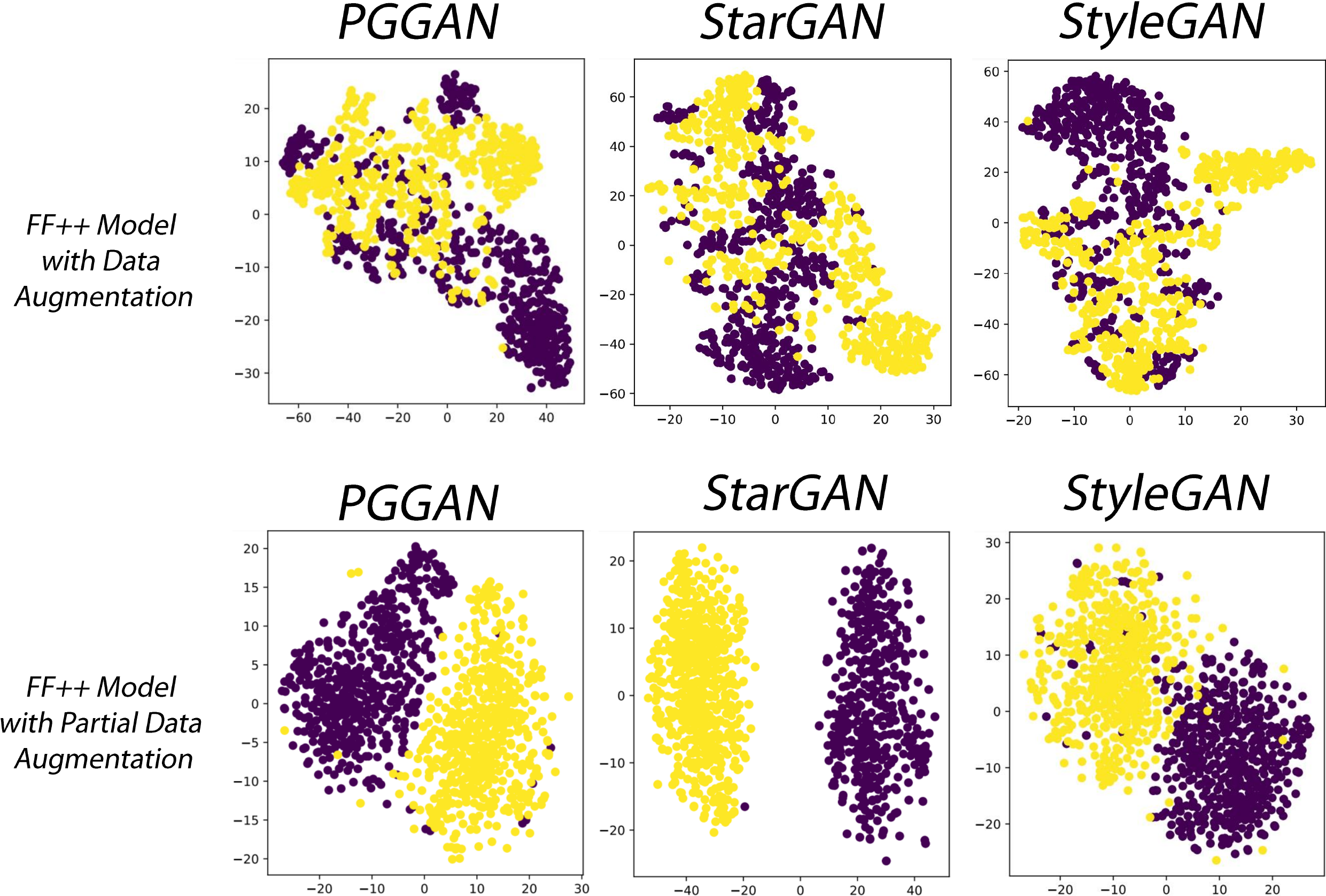}
	\caption{The t-SNE visualization of features before classifier. The models are on the left and the test datasets are above the right three subfigures.}
	\label{fig:t-sne_embedding_with_aug_and_partial_aug}
\end{figure}

Furthermore, the FaceForensics++ model with data augmentation has a similar conclusion as that without data augmentation. As shown in the first row of Fig.~\ref{fig:t-sne_embedding_with_aug_and_partial_aug}, the features of real images are entangled with fake images. Compared with that, in the second row of Fig.~\ref{fig:t-sne_embedding_with_aug_and_partial_aug}, the model is the FaceForensics++ model with partial data augmentation. We can find that the real and fake images features of PGGAN, StarGAN, and StyleGAN are clearly distinct, even a simple linear classifier can distinguish them, which violates the low accuracy in Table~\ref{tab:augmentation_general_test_diff_DeepFake_category}. It is obvious that the classifier of the FaceForensics++ partial data augmentation model is not suitable for distinguishing PGGAN, StarGAN, and StyleGAN. That is to say, the universality of the classifier is not enough for cross-method detection. 

To deal with this problem, we propose to use clustering with a single sample, which does not include the classifier and only uses the features before the classifier. To be specific, we use k-means \cite{hartigan1979algorithm} to cluster the dimension reduction results from t-SNE into two clusters without labels. Then we randomly choose a single sample that has a ground truth label to confirm which cluster is features of real images and which cluster is features of fake images (\ie, we just need to access one single sample with its label from the unknown target GAN). To mitigate the error of random selection, we run the method ten times and calculate the average accuracy.

As shown in the second row of Table~\ref{tab:augmentation_general_test_diff_DeepFake_category}, we can find that the accuracy (\ie, the third value in each cell) of the FaceForensics++ partial data augmentation model on StarGAN, StyleGAN, and PGGAN significantly increase, some even doubled.
As shown in the second column of Table~\ref{tab:augmentation_general_test_diff_DeepFake_category}, we can find that the accuracy (\ie, the third value in each cell) of StarGAN, StyleGAN, and PGGAN partial data augmentation model on the FaceForensics++ dataset also significantly increase.



\subsection{Implementation Details and Analysis of \emph{FakeLocator}}\label{Implementation}
\subsubsection{Gray-scale \emph{vs.} Binary Fakeness Map}\label{Data Pre-process}
In the existing GAN-based face generation methods, even the best of them changes most of the pixels in the image when manipulating a property of the face. If we set the values of manipulated pixels as 1 while unmodified pixels as 0, then the fakeness prediction map does not catch the emphasis of the change in the figure. 

To solve the problem, other localization methods \cite{li2020face,dang2020detection,songsri2019complement} use the \textbf{binary fakeness map}. They produce difference maps between real images and fake images and use a threshold to generate binary fakeness maps from the difference maps. The pixel values larger than the threshold become 1 while the others become 0. However, the setting of a threshold is a big flop. Firstly, all the information about fake regions less than the threshold is omitted. Secondly, the threshold is usually a fixed value defined by experience, which may not work in all cases. Finally, the values of manipulated pixels in the fake image are all less than the threshold if the fake image is slightly manipulated. The binary fakeness map thus turns out to be entirely black and far from the truth. 

\begin{figure}[tbp]
	\centering 
	\includegraphics[width=\columnwidth]{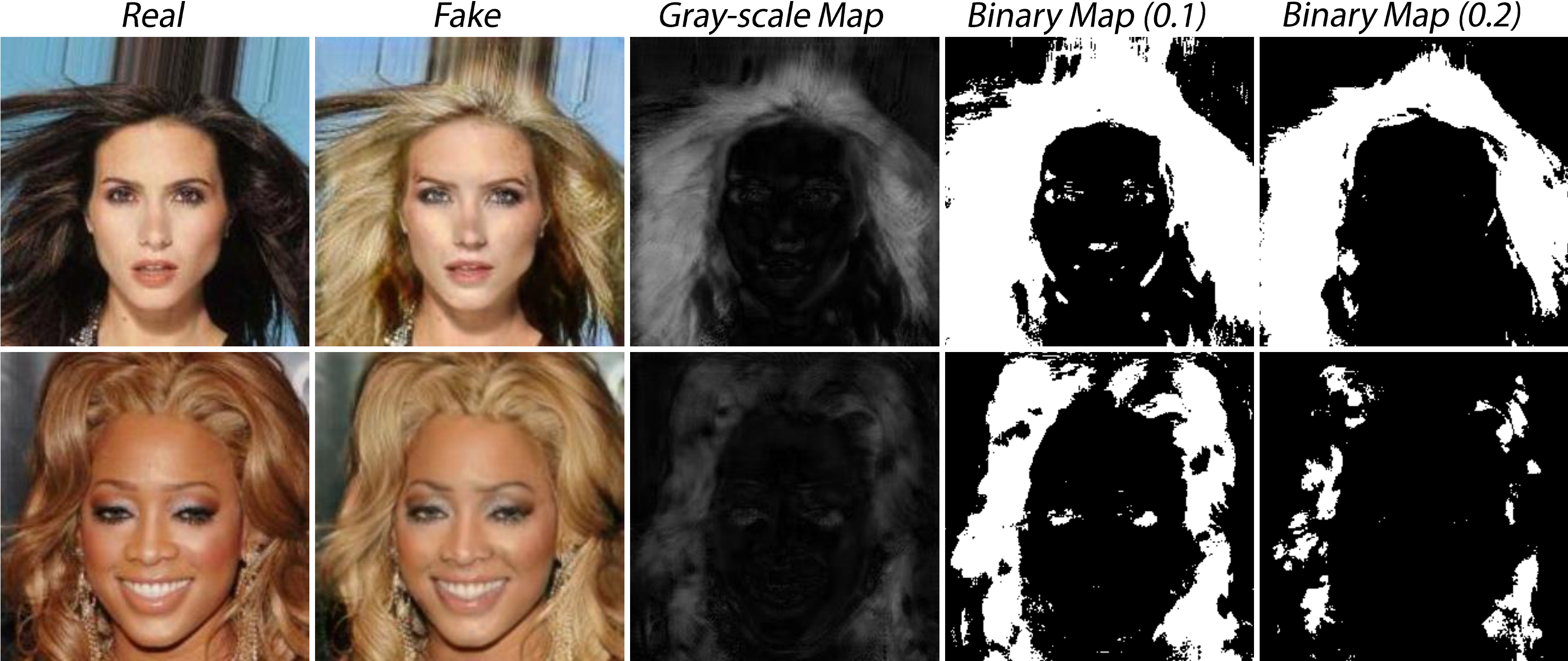}
	\caption{The images from left to right: real images, fake images generated by modifying with StarGAN of \emph{Blond hair} property, the gray-scale fakeness map, the binary fakeness map generated by threshold 0.1, the binary fakeness map generated by threshold 0.2.}
	\label{fig:binary_gray-scale_comparison}
\end{figure}

Here we demonstrate the defect of the binary fakeness map clearly. As shown in Fig. \ref{fig:binary_gray-scale_comparison}, we show the influence of different thresholds on the binary fakeness map. The modified facial property is \emph{Blond hair}. In the first row, according to the same difference map (\ie, the gray-scale fakeness map used by us), the binary map with threshold 0.1 not only points out the difference in hair but also points out the difference on eye and mouth, which is imprecise. It can be observed that with a 0.2 threshold we can get a basically satisfied binary fakeness map. However, the gray-scale fakeness map has already reflected the difference between the variation of hair area and non-modified area. There is no need to use a threshold. Moreover, as shown in the second row, the binary map of threshold 0.1 is better than that of threshold 0.2, which is different from the situation of the first row. Just for two images with the same properties modified, we need different thresholds. How to choose a suitable threshold for other different facial properties is a more serious problem. To summarize, using the binary map to represent the fakeness map in the DeepFake localization task is too subjective, which is not as flexible as the gray-scale map.

Therefore, we discard the threshold when producing our fakeness map. The training samples consist of two parts: input image and \textbf{gray-scale fakeness ground truth map}. The input images fall into two categories, real images and fake images. For a training sample that regards the real image as the input image, the corresponding fakeness map is a gray-scale ground truth map with all the pixel values equal to 0. This means that the current input image has no fake texture. On the other hand, if the input image is a fake image, the gray-scale fakeness ground truth map is obtained by the following procedure.

\begin{figure}[tbp]
	\centering 
	\includegraphics[width=\columnwidth]{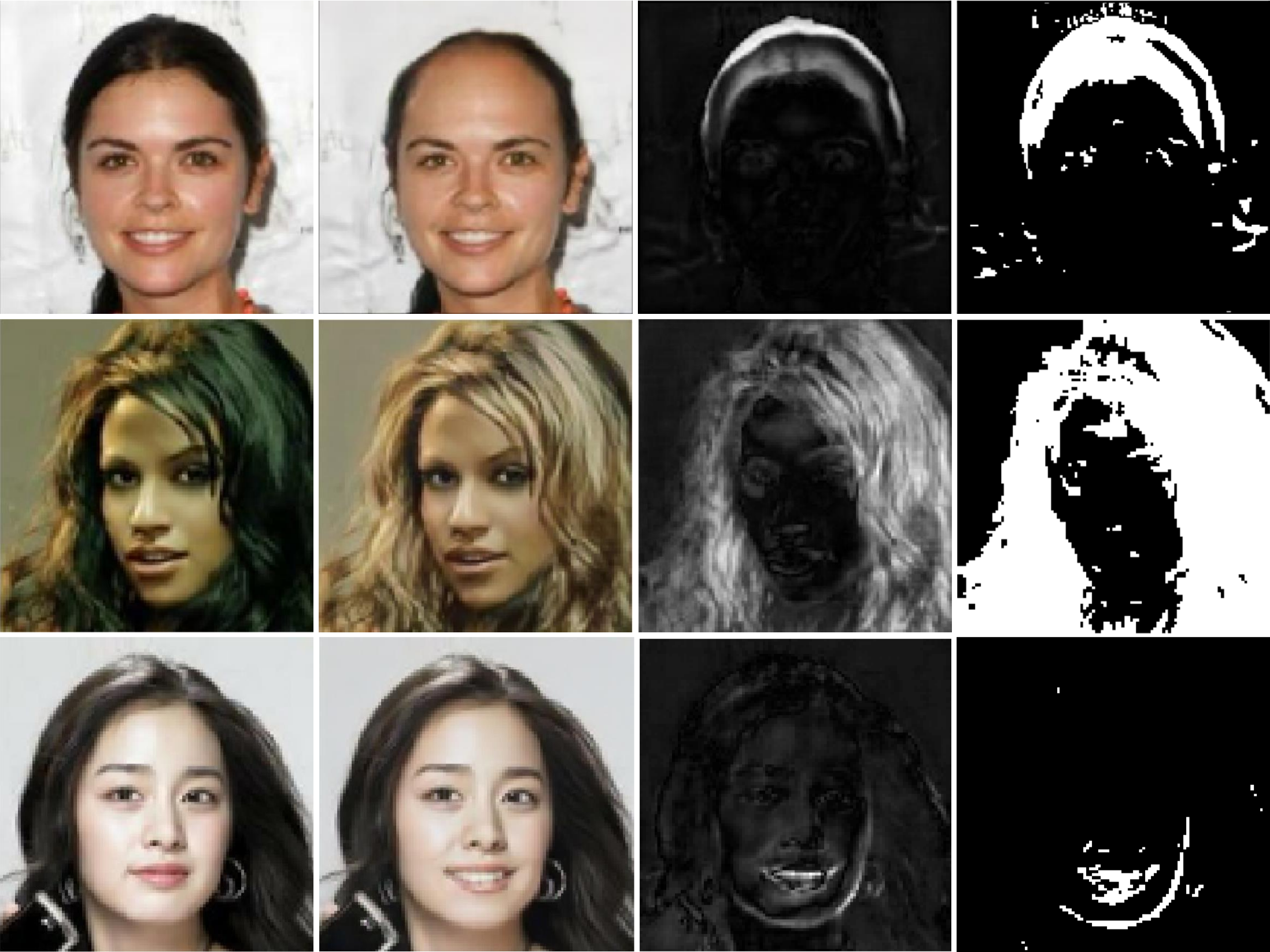}
	\caption{From left to right: real image, fake image, gray-scale fakeness ground truth map, and binary fakeness ground truth map. The modified properties from top to bottom are \emph{Bald}, \emph{Blond hair}, and \emph{Smile}.}
	\label{fig:fake_image_and_segmentation_map}
\end{figure}

As shown in Fig.~\ref{fig:fake_image_and_segmentation_map}, these images in turn are real image $\mathbf{X}$, fake image $\mathbf{X^{'}}$, gray-scale fakeness ground truth map $\mathbf{M}_{\mathrm{GT\_G}}$ and binary fakeness ground truth map $\mathbf{M}_{\mathrm{GT\_B}}$. 
$\mathbf{X}$, $\mathbf{X^{'}}$ $\in\mathbb{R}^{H\times W\times 3}$ and $\mathbf{M}_{\mathrm{GT\_G}}$, $\mathbf{M}_{\mathrm{GT\_B}}$ $\in\mathbb{R}^{H\times W\times 1}$, where $H$, $W$ are height and width of the these images. 
The fake image is produced by adding property $p$ to the real image. 
\begin{equation}
    \setlength\abovedisplayskip{4pt}
    \setlength\belowdisplayskip{4pt}
    \mathbf{X^{'}}= \mathcal{G}_{\mathrm{dec},S}( \mathcal{G}_{\mathrm{enc},S}(\mathbf{X}),p).
\end{equation}
$\mathcal{G}_{\mathrm{dec},S}$ and $ \mathcal{G}_{\mathrm{enc},S}$ are the decoder and encoder of the method $S$, $S\in $ \{all the partial face manipulation GANs\}. To achieve the gray-scale fakeness ground truth map, there are three steps as follows. (1) calculate the pixel difference between the real image and the fake image. (2) take the absolute value of the result and turn it into a gray-scale map. (3) divide each pixel by 255. Then each pixel value falls in the range of [0,1]. Eq.~\eqref{RSM} shows the formula, in which $\mathbf{X}_{i,j,k}$, $\mathbf{X}_{i,j,k}'$ $(1\le i\le H, 1\le j \le W, 1\le k \le 3, 0\le \mathbf{X}_{i,j,k},\mathbf{X}_{i,j,k}'\le 255)$ are the value of a channel of a pixel of $\mathbf{X}$ and $\mathbf{X^{'}}$ respectively.  $\mathbf{M}_{\mathrm{GT\_G_{i,j}}}$, $\mathbf{M}_{\mathrm{GT\_B_{i,j}}}$ $(1\le i\le H, 1\le j\le W, 0\le \mathbf{M}_{\mathrm{GT\_G_{i,j}}}\le 1, \mathbf{M}_{\mathrm{GT\_B_{i,j}}}\in \{0,1\})$ are the value of a pixel of $\mathbf{M}_{\mathrm{GT\_G}}$ and $\mathbf{M}_{\mathrm{GT\_B}}$, respectively. $\mathrm{Gray}(\cdot)$ is a function that converts an RGB pixel to a gray-scale pixel.
\begin{equation}
    \setlength\abovedisplayskip{4pt}
    \setlength\belowdisplayskip{4pt}
    \mathbf{M}_{\mathrm{GT\_G_{i,j}}} =  \mathrm{Gray}  \Big(  | \mathbf{X}_{i,j,k} - \mathbf{X}_{i,j,k}'  |  \Big) /255.
    \label{RSM}
\end{equation}

For the entire face synthesis, the fakeness ground truth map is a gray-scale map with all the pixel values equal to 1. The gray-scale fakeness ground truth map clearly depicts the difference between real and fake images and highlights the regions with large differences.

\subsubsection{Loss Function for Localization}\label{Loss function}
For the gray-scale fakeness map, we have tried four loss functions. Two of them are $\mathcal{L}_{1}$ loss and $\mathcal{L}_{2}$ loss, which is commonly used in regression problems. The other two are Focal loss \cite{lin2017focal} and Dice loss \cite{sudre2017generalised}, which is commonly used in traditional segmentation problems. In the experiment, we find that $\mathcal{L}_{1}$ loss and $\mathcal{L}_{2}$ loss are better choices than the others. This means that for gray-scale fakeness maps, regression losses are better than traditional segmentation losses. The comparison is shown in Table \ref{tab:loss_comparison}. The formulas of $\mathcal{L}_{1}$ and $\mathcal{L}_{2}$ are shown in Eq.~\eqref{L1Loss} and \eqref{L2Loss}. $\mathbf{M}_{\mathrm{Pred\_G_{i,j}}}$ and $\mathbf{M}_{\mathrm{GT\_G_{i,j}}}$ $(1\le i\le H, 1\le j \le W)$ represent the pixel in the gray-scale fakeness prediction map and of the gray-scale fakeness ground truth map respectively, where $H$, $W$ are the height and width of the maps. 
\begin{align}
    \setlength\abovedisplayskip{3pt}
    \setlength\belowdisplayskip{3pt}
    \mathcal{L}_{1} &= \frac{1}{n} \sum  | \mathbf{M}_{\mathrm{Pred\_G_{i,j}}} - \mathbf{M}_{\mathrm{GT\_G_{i,j}}}  |. \label{L1Loss}\\
    \mathcal{L}_{2} &= \frac{1}{n} \sum  |\mathbf{M}_{\mathrm{Pred\_G_{i,j}}} - \mathbf{M}_{\mathrm{GT\_G_{i,j}}}  |^{2}.
    \label{L2Loss}
\end{align}

The formula of Focal loss is shown below in Eq.~\eqref{Pt} and \eqref{FocalLoss}. Assume $p \in$ [0,1] is the model’s estimated probability for the class with label $y$ = 1.
\begin{equation}
    p_{t}=\left\{
    \begin{array}{rcl}
    p       &      & {y = 1}\\
    1-p     &      & \mathrm{otherwise}
    \end{array} \right..
    \label{Pt}
\end{equation}
\begin{equation}
    \mathcal{L}_{\mathrm{Focal}}(p_{t})= -\alpha_{t}(1-p{t})^{\gamma}\log(p_{t}).
    \label{FocalLoss}
\end{equation}
$\alpha_{t}$ is an equilibrium factor. The focusing parameter $\gamma$
smoothly adjusts the rate at which easy examples are down-weighted.

The formulation of Dice loss is shown below in Eq.\eqref{DiceLoss}. Function $\mathrm{PS}(\cdot)$ represents the point set of any fakeness map. $\mathrm{PS}(\mathbf{M}_{\mathrm{GT}})$ and $\mathrm{PS}(\mathbf{M}_{\mathrm{Pred}})$ represent the point set of fakeness ground truth map and fakeness prediction map respectively.
\begin{equation}
    \mathcal{L}_{\mathrm{Dice}} = \frac {2|\mathrm{PS}(\mathbf{M}_{\mathrm{GT}}) \cap \mathrm{PS}(\mathbf{M}_{\mathrm{Pred}})|} {|\mathrm{PS}(\mathbf{M}_{\mathrm{GT}})|+|\mathrm{PS}(\mathbf{M}_{\mathrm{Pred}})|}.
    \label{DiceLoss}
\end{equation}
%
For training the classification network (\ie, $\mathcal{C}(\cdot)$) in Eq.~\eqref{cls_pred}, we add the cross-entropy loss and jointly train it with the encoder-decoder architecture. We also discuss the influence of this additional module in Table \ref{tab:loss_comparison}.

\section{Experimental Result}\label{Experiment}
\subsection{Experimental Setup} \label{Experimental_Setup}

\textbf{Databases.}
Our experiment benchmarks the real face databases CelebFaces Attributes (CelebA) \cite{liu2018large} and Flickr-Faces-HQ (FFHQ) \cite{karras2019style}.  CelebA contains 202,599 face images of celebrities, each annotated with 40 binary attributes. FFHQ contains 70,000 high-quality images. The database includes vastly more variation than CelebA in terms of age, ethnicity, and image background.

The real images are from CelebA and FFHQ databases. The fake images are produced by seven GAN-based face generation methods and their corresponding databases and properties, as shown in Table \ref{tab:GAN-property}. The images are all in PNG format. We use \textit{Entire} to represent the property of the images produced by entire face synthesis methods. 
\begin{table}[!htbp]
\centering
\caption{Properties of GANs. We choose the facial properties which well modified by these GANs.}
\label{tab:GAN-property}
\begin{tabular}{c|c}
\toprule
STGAN (CelebA)                                                                                                               & AttGAN (CelebA)                                                                                                     \\ \midrule
\begin{tabular}[c]{@{}c@{}}Bald, Bangs, Black hair,\\ Blond hair, Smile, Brown hair, \\ Eyeglasses, Male, Mustache\end{tabular} & \begin{tabular}[c]{@{}c@{}}Bald, Bangs, Black hair, Blond hair, \\ Brown hair, Eyeglasses, Male, Smile\end{tabular} \\ \midrule

StarGAN (CelebA)                                                                                                             & IcGAN (CelebA)                                                                                                      \\ \midrule
\begin{tabular}[c]{@{}c@{}}Black hair, Blond hair, \\ Brown hair, Gender, Age\end{tabular}                                   & \begin{tabular}[c]{@{}c@{}}Bald, Bangs, \\ Eyeglasses, Smile\end{tabular}                               \\ \midrule

StyleGAN (FFHQ)                                                                                                              & \begin{tabular}[c]{@{}c@{}}PGGAN (FFHQ),StyleGAN (FFHQ),\\ StyleGAN2 (FFHQ)\end{tabular}                            \\ \midrule

Smile, Age, Gender                                                                                                            & Entire                                                                                                              \\ \bottomrule
\end{tabular}
\end{table}
We also take FaceForensics++, UADFV, Celeb-DF, and DFFD as the DeepFake dataset. 

\textbf{Settings.}
All the experiments were run on an Ubuntu 16.04 system with an Intel(R) Xeon(R) CPU E5-2699 with 196 GB of RAM, with an NVIDIA Tesla V100 GPU of 32G RAM. In the experiment, we train for 10 epochs with an Adam \cite{kingma2014adam} optimizer whose $\alpha$ = 0.0001, $\beta_{1}$ = 0.99, $\beta_{2}$ = 0.999, and weight decay = $10^{-7}$. The minibatch size of every training and validation dataset is 8. 

\textbf{Network architecture.}
Any encoder-decoder network is feasible in our framework. In the available networks, we choose the Deeplabv3 architecture as our backbone network. Deeplabv3 has a good performance in segmentation and PyTorch \cite{paszke2019pytorch} provides a pre-trained Deeplabv3-ResNet101 model. Deeplabv3-ResNet101 is constructed by a Deeplabv3 model with a ResNet-101 backbone. The pre-trained model has been trained on a subset of COCO train2017, on the 20 categories that are present in the Pascal VOC dataset. We fine-tune the network and use the Sigmoid function as the limiting layer. The input size supported by Deeplabv3-ResNet101 is 224 $\times$ 224, thus we resized all the input images to this size. Other encoder-decoder networks are also available. In terms of the classification network in Eq.~\eqref{cls_pred}, it contains four convolutional layers (kernel size is 3), a max-pooling layer (the kernel size is 2, the stride is 2, padding is 0), three fully connected layers and a Sigmoid layer for classification. It is just added for facilitating the calculation of the accuracy of detecting whether an image is real or fake. 
As a result, we can also compare our method with state-of-the-art DeepFake detection methods. 

\textbf{Metrics.}
We evaluate our method from two aspects, \ie, fake localization and classification. 
For the fake classification, we employ the classification accuracy (ACC) on all images, accuracy on the real images (real ACC), and accuracy on the fake images (fake ACC) as the metrics. 
We also report the equal error rate (EER) and area under the curve (AUC) for a comprehensive comparison.
Note that, we report the classification results since the accuracy results are the foundation of localization. Only with high accuracy results, is the localization result meaningful. 
For the evaluation of fake localization (\ie, the quality of fakeness prediction maps), we use cosine similarity (COSS), peak signal-to-noise ratio (PSNR) and structural similarity (SSIM), intersection over union (IoU), Dice coefficient (Dice), pixel-wise binary classification accuracy (PBCA) and inverse intersection non-containment (IINC) as the metrics. COSS is a measure of similarity between two non-zero vectors of an inner product space that measures the cosine of the angle between them. PSNR is the most commonly used measurement for the reconstruction quality of lossy compression. SSIM is used for measuring the similarity between two images. IoU is the most popular evaluation metric for segmentation tasks. Dice is a spatial overlap index and a reproducibility validation metric. PBCA treats each pixel as an independent sample to measure classification accuracy. IINC is a more robust metric for comparing the prediction maps proposed by \cite{dang2020detection}. To enable the calculation of IoU, Dice, PBCA and IINC, we transform the gray-scale fakeness maps to binary fakeness maps. The pixels less than 0.1 become 0 while the others become 1. The threshold value 0.1 is referred to the choice of \cite{dang2020detection}, which has been verified to be a good choice. ACC, AUC, COSS, PSNR, SSIM, IoU, Dice, PBCA metrics are better if a higher value is provided while a lower value is better for EER and IINC. The value ranges of ACC, EER, AUC, COSS, SSIM, IoU, Dice, PBCA and IINC are all in [0,1]. When compared with baseline methods, we use the metrics. For ablation study of our method, we select COSS, PSNR, SSIM as the representative localization metric for that the IoU, Dice, PBCA, and IINC are likely to change with different thresholds chosen by us.

\subsection{Results and Analysis}
Experiments evaluate the effectiveness, robustness, and universality of our method. 

\subsubsection{Comparison with detection methods}
Before evaluating the localization performance of our method, we add extra experiments to verify the effectiveness of the detection performance of our method. After all, the ability of localization is only meaningful if the detection is good.
We take state-of-the-art methods (CNNDetection \cite{wang2020cnn} and DCTA \cite{frank2020leveraging}) as the baselines. We use the popular datasets UADFV \cite{li2018ictu} and Celeb-DF \cite{li2020celeb} as the DeepFake dataset. For UADFV, all three methods are trained with 28,000 images and tested on 2,000 images. For Celeb-DF, all three methods are trained with 280,000 images and tested on 20,000 images. As shown in Table \ref{tab:detection_comparison}, we can find that our method has a comparable result with CNNDetection and does better than DCTA.

\begin{table}[tbp]
\footnotesize
\centering
\caption{Comparison with detection methods. We compare our method with CNNDetection \cite{wang2020cnn} and DCTA \cite{frank2020leveraging} on UADFV \cite{li2018ictu}, and Celeb-DF \cite{li2020celeb}. Here shows the accuracy result of each method.}
\setlength{\tabcolsep}{3pt}
\label{tab:detection_comparison}
\begin{tabular}{c|ccc|ccc}
\toprule 
 & \multicolumn{3}{c|}{UADFV} & \multicolumn{3}{c}{Celeb-DF}\tabularnewline
\midrule 
 & Ours & CNNDetection & DCTA & Ours & CNNDetection & DCTA\tabularnewline
ACC & 1.0 & 1.0 & 0.8560 & 0.9304 & 0.9992 & 0.5800\tabularnewline
real ACC & 1.0 & 1.0 & - & 0.8560 & 0.9993 & -\tabularnewline
fake ACC & 1.0 & 1.0 & - & 0.9977 & 0.9991 & -\tabularnewline
\bottomrule 
\end{tabular}
\end{table}

\subsubsection{Comparison with baseline}
There are two methods selected by us as the \textbf{Baseline}: \cite{dang2020detection} and \cite{patchforensics}. As the method of \cite{dang2020detection} has an underlying theoretical flaw (low-resolution fakeness map) in localization, we just simply use STGAN with specific property to show the huge gap between their performance and ours. In Table \ref{tab:loss_comparison}, we show a comparison of our method and the \textbf{Baseline} \cite{dang2020detection}. The table also shows the comparison of results with different loss function strategies. Here we choose \emph{Bald} of STGAN as the property of fake images. In the first column, Baseline represents the result of \cite{dang2020detection}. We enlarge its binary fakeness prediction maps to the size of 224 $\times$ 224 before calculating the metrics. The label \textit{no cl} means the model that does not have a classifier. It is obvious that, compared to \textbf{Baseline}, the gray-scale fakeness prediction maps predicted by our method not only have a higher resolution, but also a better performance (several times higher). Moreover, the performance is distinctly higher on all the other properties and GAN-based face generation methods.

For \textbf{Baseline} \cite{patchforensics}, we use DeepFake dataset DFFD, which is collected by \cite{dang2020detection} to show the comparison. DFFD contains images from FaceForensics++, faceapp, StarGAN, PGGAN, and StyleGAN. 
Both our method and \cite{patchforensics} are trained with 12,618/ 39,914/ 43,698/ 39,998 images of faceapp/ PGGAN/ StarGAN/ StyleGAN as the training dataset. The test dataset for them are all 9,000 images. For identity swap videos in FaceForensics++, we use 180,000 images as the training dataset and 18,000 images as the test dataset. As shown in Table \ref{tab:localization_comparison_DFFD}, we can find that our method has better performance on almost all the metrics than \cite{patchforensics} among all the four datasets.

\begin{table*}[tbp]
\centering
\setlength{\tabcolsep}{5.6pt}
\caption{Loss Function Comparison. Here we compare the performance of different loss strategies. We choose \emph{Bald} of STGAN as the property of fake images. In the first column, Baseline represents the result of \cite{dang2020detection}. The label \textit{no cl} means the model that does not have a classifier.}
\label{tab:loss_comparison}
\begin{tabular}{lcccccccccccc}
\toprule 
 & ACC & real ACC & fake ACC & AUC & EER & COSS & PSNR & SSIM & Dice & IoU & PBCA & IINC\tabularnewline
\midrule 
\rowcolor{lightgreen}
\textbf{Baseline \cite{dang2020detection}} & 0.9975 & 0.995 & 1.000 &  0.9975 & 0.0050 & 0.6230 & 6.214 & 0.2178 & 0.1415 & 0.1432 & 0.3211 & 0.4301\tabularnewline
$L_1$ Loss (\textit{no cl}) & \textbackslash{} & \textbackslash{} & \textbackslash{} & \textbackslash{} & \textbackslash{} & {\color[HTML]{FE0000} 0.9271} & 22.54 & 0.7533 & 0.1975 & 0.5359 & 0.8985 & {\color[HTML]{FE0000} 0.1898}\tabularnewline
$L_1$ Loss & 0.9945 & 0.998 & 0.991 & 0.9998 & 0.0077 & 0.8813 & {\color[HTML]{FE0000} 22.89} & {\color[HTML]{FE0000} 0.7876} & 0.1899 & 0.5517 & 0.9272 & 0.1996\tabularnewline
$L_2$ Loss & 0.9935 & 0.999 & 0.988 & 0.9997 & 0.0072 & 0.8916 & 22.36 & 0.7452 & 0.1891 & 0.5371 & 0.9078 & 0.2010\tabularnewline
Dice Loss & 0.9950 & 1.000 & 0.990 & 0.9989 & 0.0080 & 0.7645 & 13.04 & 0.3581 & {\color[HTML]{FE0000} 0.2904} & {\color[HTML]{FE0000} 0.5573} & {\color[HTML]{FE0000} 0.9339} & 0.1977\tabularnewline
Focal Loss & 0.9920 & 0.996 & 0.988 & 0.9995 & 0.0095 & 0.4048 & 20.74 & 0.3287 & 0.0708 & 0.1369 & 0.8872 & 0.3357\tabularnewline
\bottomrule 
\end{tabular}
\end{table*}

\begin{table*}[tbp]
\centering
\caption{Comparison with localization method. Here we compare the localization and detection performance of our method with \cite{dang2020detection} and \cite{patchforensics}. In the first column are the different DeepFake methods in dataset DFFD and dataset FaceForensics++.}
\setlength{\tabcolsep}{5pt}
\label{tab:localization_comparison_DFFD}
\resizebox{1\linewidth}{!}{
\begin{tabular}{lccccccccccccc}
\toprule 
 & \multirow{1}{*}{} & ACC & \multirow{1}{*}{real ACC} & fake ACC & \multirow{1}{*}{AUC} & EER & \multirow{1}{*}{COSS} & PSNR & SSIM & Dice & IoU & PBCA & IINC\tabularnewline
\midrule 
\multirow{3}{*}{DFFD (faceapp)} & Ours & 0.9998 & 1 & 0.9997 & 1 & 0.2355 & 0.7250 & 23.11 & 0.6550 & 0.0974 & 0.2384 & 0.8092 & 0.4036\tabularnewline
& \textbf{Baseline} \cite{dang2020detection} & 1 & 1 & 1 & 1 & 0 & 0.5538 & 6.874 & 0.1848 & 0.0937 & 0.0448 & 0.2964 & 0.4499\tabularnewline
& \textbf{Baseline} \cite{patchforensics} & 0.9983 & 0.9995 & 0.9971 & 0.9999 & 0.0020 & 0.5985 & 0.7197 & 0.0570 & 0.0779 & 0.0996 & 0.1017 & 0.4510\tabularnewline
\midrule 
\multirow{3}{*}{DFFD (StarGAN)} & Ours & 0.9997 & 1 & 0.9995 & 1 & 0.0086 & 0.8373 & 19.39 & 0.6433 & 0.1822 & 0.4953 & 0.6749 & 0.2551\tabularnewline
& \textbf{Baseline} \cite{dang2020detection} & 1 & 1 & 1 & 1 & 0 & 0.5308 & 16.23 & 0.5541 & 0.1137 & 0.0345 & 0.5733 & 0.5141\tabularnewline
& \textbf{Baseline} \cite{patchforensics} & 1 & 1 & 1 & 1 & 0 & 0.7462 & 1.019 & 0.1453 & 0.2033 & 0.3887 & 0.3887 & 0.3056\tabularnewline
\midrule 
\multirow{3}{*}{DFFD (PGGAN)}  & Ours& 1 & 1 & 1 & 1 & 0 & 1 & 142.8 & 0.9999 & 0.9999 & 1 & 1 & 0\tabularnewline
& \textbf{Baseline} \cite{dang2020detection} & 0.9997 & 0.9995 & 1 & 0.9997 & 0.0004 & 0.9231 & 4.656 & 0.5358 & 0.6148 & 0.9809 & 0.9809 & 0.0095\tabularnewline
& \textbf{Baseline} \cite{patchforensics} & 1 & 1 & 1 & 1 & 0 & 0.9984 & 28.86 & 0.9745 & 0.9957 & 0.9993 & 0.9993 & 0.0003\tabularnewline
\midrule 
\multirow{3}{*}{DFFD (StyleGAN)}  & Ours& 0.9998 & 1 & 0.9997 & 0.9999 & 0.0024 & 0.9998 & 127.3 & 0.9997 & 0.9997 & 0.9998 & 0.9998 & 0.0001\tabularnewline
& \textbf{Baseline} \cite{dang2020detection} & 1 & 1 & 1 & 1 & 0 & 0.7927 & 3.286 & 0.3186 & 0.5472 & 0.8052 & 0.8052 & 0.0973\tabularnewline
& \textbf{Baseline} \cite{patchforensics} & 0.9994 & 1 & 0.9988 & 0.9999 & 0.0008 & 0.9868 & 19.01 & 0.8779 & 0.9710 & 0.9921 & 0.9921 & 0.0039\tabularnewline
\midrule 
\multirow{3}{*}{FaceForensics++} & Ours & 0.9398 & 0.9275 & 0.9521 & 0.9846 & 0.0576 & 0.7529 & 26.94 & 0.7691 & 0.0811 & 0.3097 & 0.8687 & 0.3143\tabularnewline
& \textbf{Baseline} \cite{dang2020detection} & 0.8714 & 0.8042 & 0.9386 & 0.8714 & 0.1725 & 5.539 & 0.1877 & 0.5739 & 0.0863 & 0.1174 & 0.3391 & 0.4194\tabularnewline
& \textbf{Baseline} \cite{patchforensics} & 0.4909 & 0.4667 & 0.5150 & 0.4659 & 0.5332 & 0.5208 & 6.717 & 0.1017 & 0.0689 & 0.1056 & 0.1230 & 0.4432\tabularnewline
\bottomrule 
\end{tabular}}
\end{table*}

\begin{table}[tbp]
\centering
\setlength{\tabcolsep}{5pt}
\caption{Universal Test. Here are two different test pairs. The model trained by STGAN (Bald) tests properties of AttGAN and the model trained by STGAN (Smile) tests properties of StyleGAN.}
\label{tab:general_test}
\resizebox{\linewidth}{!}{
\begin{tabular}{c|
>{\columncolor{lightgreen}}c  c c c|
>{\columncolor{lightgreen}}c  c c c}
\toprule
& \cellcolor{lightgreen}& \multicolumn{3}{c|}{AttGAN}& \cellcolor{lightgreen}& \multicolumn{3}{c}{StyleGAN} \\  
 & \multirow{-2}{*}{\cellcolor{lightgreen}\begin{tabular}[c]{@{}c@{}}STGAN\\Bald\end{tabular}} & Bald   & Black hair & Eyeglasses & \multirow{-2}{*}{\cellcolor{lightgreen}\begin{tabular}[c]{@{}c@{}}STGAN\\Smile\end{tabular}} & Smile    & Age      & Gender  \\ \midrule
ACC                 & 0.994                                                                                          & 0.973  & 0.513      & 0.519      & 1.000                                                                                         & 0.588    & 0.593    & 0.588   \\ 
PSNR               & 22.83                                                                                          & 21.46  & 22.84      & 23.52      & 35.49                                                                                           & 20.85    & 18.59    & 16.50   \\ 
SSIM               & 0.7823                                                                                         & 0.6688 & 0.3966     & 0.4168     & 0.9218                                                                                          & 0.5571   & 0.4958   & 0.4301  \\ 
COSS             & 0.8887                                                                                         & 0.8141 & 0.6932     & 0.6950     & 0.8844                                                                                          & 0.6176   & 0.3141   & 0.6470  \\ \bottomrule

\end{tabular}}
\end{table}

\begin{table}[tbp]
\centering
\setlength{\tabcolsep}{1pt}
\caption{Ablation study of the functionality of partial augmentation on the universal test. The table mainly shows the detection performance of the model trained from AttGAN (Bald) testing on STGAN.}
\label{tab:ablation_study_general_test}
\resizebox{\linewidth}{!}{
\begin{tabular}{c| >{\columncolor{lightgreen}}ccccc}
\toprule 
 & AttGAN & STGAN & STGAN & STGAN & STGAN\tabularnewline
 & no aug & no aug & real(1.0)fake(1.0) & real(1.0)fake(0.0) & real(0.0)fake(1.0)\tabularnewline
\midrule 
ACC & 0.998 & 0.7315 & 0.641 & 0.9535 & 0.521\tabularnewline
real ACC & 0.998 & 0.9980 & 1.000 & 0.9210 & 1.000\tabularnewline
fake ACC & 0.998 & 0.4650 & 0.282 & 0.9860 & 0.042\tabularnewline
\bottomrule 
\end{tabular}}
\end{table}

\begin{table*}[tbp]
\centering
\caption{The effect of partial data augmentation on improving cross-method universality. In the first row are the facial properties used in the dataset of the below columns. There are six pairs of experiments. For each of them, the first column records the value of the model testing on the test dataset of itself. The second/third column records the value of the model testing on the other test dataset without/with partial data augmentation, where ``real($p_{real}$)fake($p_{fake}$)'' represents the probability of a real/fake image being augmented in the training procedure.}
\label{tab:augmentation_general_test}
\resizebox{1\linewidth}{!}{
\begin{tabular}{cccc|ccc|ccc}
\toprule 
 & \multicolumn{3}{c|}{Bald} & \multicolumn{3}{c|}{Blond hair} & \multicolumn{3}{c}{Brown hair}\tabularnewline 
 \midrule 
 & AttGAN & STGAN & STGAN & AttGAN & STGAN & STGAN & AttGAN & STGAN & STGAN\tabularnewline
 & no aug & no aug & real(1.0)fake(0.0) & no aug & no aug & real(1.0)fake(0.0) & no aug & no aug & real(1.0)fake(0.0)\tabularnewline
\midrule 
ACC & 0.998 & 0.7315 & 0.9535 & 1.0 & 0.5 & 0.5015 & 1.0 & 0.5 & 0.5015\tabularnewline
real ACC & 0.998 & 0.9980 & 0.9210 & 1.0 & 1.0 & 1.0000 & 1.0 & 1.0 & 0.9980\tabularnewline
fake ACC & 0.998 & 0.4650 & 0.9860 & 1.0 & 0.0 & 0.0030 & 1.0 & 0.0 & 0.0050\tabularnewline
\midrule 
 & STGAN & AttGAN & AttGAN & STGAN & AttGAN & AttGAN & STGAN & AttGAN & AttGAN\tabularnewline
 & no aug & no aug & real(1.0)fake(0.0) & no aug & no aug & real(1.0)fake(0.0) & no aug & no aug & real(1.0)fake(0.0)\tabularnewline
\midrule 
ACC & 0.993 & 0.9595 & 0.9875 & 0.9975 & 0.796 & 0.850 & 0.9995 & 0.725 & 0.8145\tabularnewline
real ACC & 0.998 & 0.9980 & 0.9970 & 1.0000 & 1.000 & 0.775 & 0.9990 & 0.999 & 0.9760\tabularnewline
fake ACC & 0.988 & 0.9210 & 0.9780 & 0.9950 & 0.592 & 0.925 & 1.0000 & 0.451 & 0.6530\tabularnewline
\bottomrule 
\end{tabular}}
\end{table*}

\begin{table*}[tbp]
\centering
\caption{Performance of The Universal Model. We train a model with many single-face-property fake images. We select STGAN, StyleGAN, AttGAN, StarGAN, IcGAN, PGGAN, StyleGAN2, and all their properties. Each category $C_{i,j} (i\in \mathrm{GANs}$, $j\in \mathrm{Properties~of~GANs}(i))$ supports 2,000 fake images into the training dataset. There are a total of 64,000 fake images and 64,000 real images randomly selected in CelebA and FFHQ.}
\label{tab:ultimate_model}
\begin{tabular}{c c c c c c c c c c}
\toprule

{} & \multicolumn{6}{>{\columncolor{lightgreen}}c}{STGAN} & \multicolumn{3}{>{\columncolor{lightyellow}}c}{StyleGAN}  \\ 
                  & Smile   & Bangs   & Bald  &Mustache & Eyeglasses &  Black hair & Age & Smile  & Entire \\ \midrule
ACC                & 0.9985  & 0.9985  & \textbf{0.9985} & 0.9935& 0.9970 & 0.9985& 0.9985  & 0.9990  & 0.9210     \\ 
PSNR              & 34.72   & 33.07   & \textbf{27.51} &36.59 & 33.23 & 31.05 & 20.71    & 22.68  & 55.08       \\ 
SSIM              & 0.9086  & 0.9033  & \textbf{0.8572} &0.8829  & 0.8663 & 0.9029 & 0.6159  & 0.6625 & 0.8268      \\ 
COSS            & 0.8634  & 0.8850  & \textbf{0.9014}& 0.9152 & 0.9093 & 0.8817 & 0.7479  & 0.7365 & 0.9174    \\ \midrule

{} &\multicolumn{3}{>{\columncolor{lightgreen}}c}{StarGAN} &\multicolumn{3}{>{\columncolor{lightyellow}}c}{AttGAN} & \cellcolor{lightgreen}IcGAN & \cellcolor{lightyellow} PGGAN & \cellcolor{lightgreen} StyleGAN2  \\  
& Age    & Gender & Black hair & Bald   & Smile &  Eyeglasses  &Bald & Entire &Entire \\ \midrule

ACC & 0.9990  & 0.9990      & 0.9990  & 0.9985 & 0.9985     & 0.9985 &0.9995 & 0.9775   &  0.8550 \\ 
PSNR & 26.38  & 26.71      & 25.25  & 25.31  & 29.49      & 28.19 &17.57  & 48.70  &72.00 \\ 
SSIM & 0.7929 & 0.7952     & 0.7911 & 0.7839 & 0.8736     & 0.8207 & 0.6002 & 0.8945  &0.9262 \\ 
COSS & 0.8494 & 0.8416     & 0.8780 & 0.8883 & 0.8277  &0.8313   & 0.8352 & 0.9972   &0.9678  \\ \bottomrule
\end{tabular}
\end{table*}

\subsubsection{Comparison of loss functions}
Using the same loss function, Deeplabv3 with/without classifier achieves a similar value on metrics. Thus the classifier used to calculate ACC does not affect the fakeness prediction map. The highest value of PSNR and SSIM is achieved by $L_1$ loss. We can conclude that the loss functions outstanding in regression problems are better for the gray-scale fakeness map. In the following experiments, we choose $L_1$ loss as the default loss function. Although Dice loss has better performance on Dice, IoU and PBCA, we don't take it as the default loss function for that the values of these metrics are strongly related to the threshold, which is unstable.

\subsubsection{Generality of our approach with different GANs and facial properties}\label{imperfection_of_universality}
In the experiment, we find that our method is effective for all the GANs and their specified facial properties. Hence there is a further question. Is the model trained on one GAN is effective for other GANs?

As shown in Table \ref{tab:general_test}, we demonstrate some of the test results. Here are two different test pairs. The model trained from STGAN (Bald) tests properties of AttGAN and the model trained from STGAN (Smile) tests properties of StyleGAN. The gray columns are the first column of each test pair. It records the value of the model testing on the test dataset of itself, which is used as the reference substance. For each test pair, the three columns right beside the gray column are the performance of the model on other datasets.

We can find that the model trained by one GAN and with specified property is more effective on other GANs with the same dataset and property than the GANs with different datasets or properties. In our observation from extensive experiments, the model trained by one GAN and specified property also has bad performance on the same GAN with other properties.

\subsubsection{Improve the universality of the method}
A simple idea is to train a model with many single-face-property fake images. Although there is no doubt that this method will improve the universality, we just want to see how well it is. We select STGAN, StyleGAN, AttGAN, StarGAN, IcGAN, PGGAN, StyleGAN2 and all their properties. Each category $C_{i,j} (i\in \mathrm{GANs}$, $j\in \mathrm{Properties~of~GANs}(i))$ supports 2,000  fake images into the training dataset. There are a total of 64,000  fake images and 64,000 real images randomly selected in CelebA and FFHQ. The model performs well on the test dataset, which is shown in Table~\ref{tab:ultimate_model}. The gray-scale fakeness prediction maps corresponding to these metric values are referred to in Fig.~\ref{fig:show}. In addition, even though the model uses a lot of GANs and properties for training, in the testing dataset, it only labels the location of fake regions of the modified facial properties. This is a nice phenomenon demonstrating that the model has learned to recognize the fake textures and distinguish them. This method not only improves the universality but also improves the performance. The metric value of STGAN (Bald) is significantly higher than that in Table~\ref{tab:loss_comparison}. 

\begin{table}[tbp]
\centering
\caption{Comparison of the model with and without face parsing.}
\label{tab:compare_face_parsing}
\begin{tabular}{c c c c c c c}
\toprule
\multirow{2}{*}{} & \multicolumn{2}{c}{Black hair} & \multicolumn{2}{c}{Brown hair} & \multicolumn{2}{c}{Smile} \\  
                  & no FP    & with FP    & no FP    & with FP    & no FP  & with FP \\ \midrule
                  
ACC               & 0.237         & 0.670           & 0.588         & 0.859           & 0.448       & 0.752          \\ 
PSNR              & 28.270        & 25.296          & 30.611        & 21.926          & 41.836      & 40.723       \\ 
SSIM              & 0.783         & 0.825           & 0.854         & 0.802           & 0.989       & 0.990        \\ 
COSS              & 0.855         & 0.817           & 0.876         & 0.863           & 0.803       & 0.808    \\
\bottomrule
\end{tabular}
\end{table}

\subsubsection{Improve the cross-method universality of the method}\label{cross-method_universality}
Here we also introduce the effect of our method on the same DeepFake category (\ie, attribute manipulation) with different GAN types. As shown in Table~\ref{tab:ablation_study_general_test}, we train the model on AttGAN (Bald) with/without data augmentation and test the performance of the model on STGAN (Bald). The gray column records the value of the model testing on the test dataset of AttGAN (Bald), which is used as the reference substance. In the second row, ``no aug'' means that the model is trained from a dataset without data augmentation. The format ``real($p_{real}$)fake($p_{fake}$)'' represents the model trained from an augmented dataset. The $p_{real}$ and $p_{fake}$ represent the probability of a real/fake image being augmented in the training procedure. For example, ``real(1.0)'' means that 100\% of the real images are processed by augmentation. The four columns right beside the gray column are the performance of the models on detecting STGAN (Bald). \Eg, the third column shows the result of a model trained on AttGAN (Bald) real(1.0)fake(1.0) testing on STGAN (Bald). From the table, we can find that if we augment all the real images and do nothing with fake images, the performance of the model on cross-method universality increases significantly. Both augmenting real and fake images or merely augmenting fake images reduces the performance of our method on cross-method universality. To fully demonstrate the effectiveness, as shown in Table~\ref{tab:augmentation_general_test}, we test the model with partial data augmentation on both AttGAN and STGAN with three different facial properties. All the accuracies increase when using models trained with partial data augmentation.

\subsubsection{Improve the cross-attribute universality of the method on unseen facial properties}
If the facial properties are known by us, the method mentioned above is useful. However, sometimes the fake images are modified with unseen facial properties. Here we introduce a method that improves the cross-attribute universality of the model on unseen facial properties.

\begin{figure}[tbp]
	\centering 
	\includegraphics[width=\linewidth]{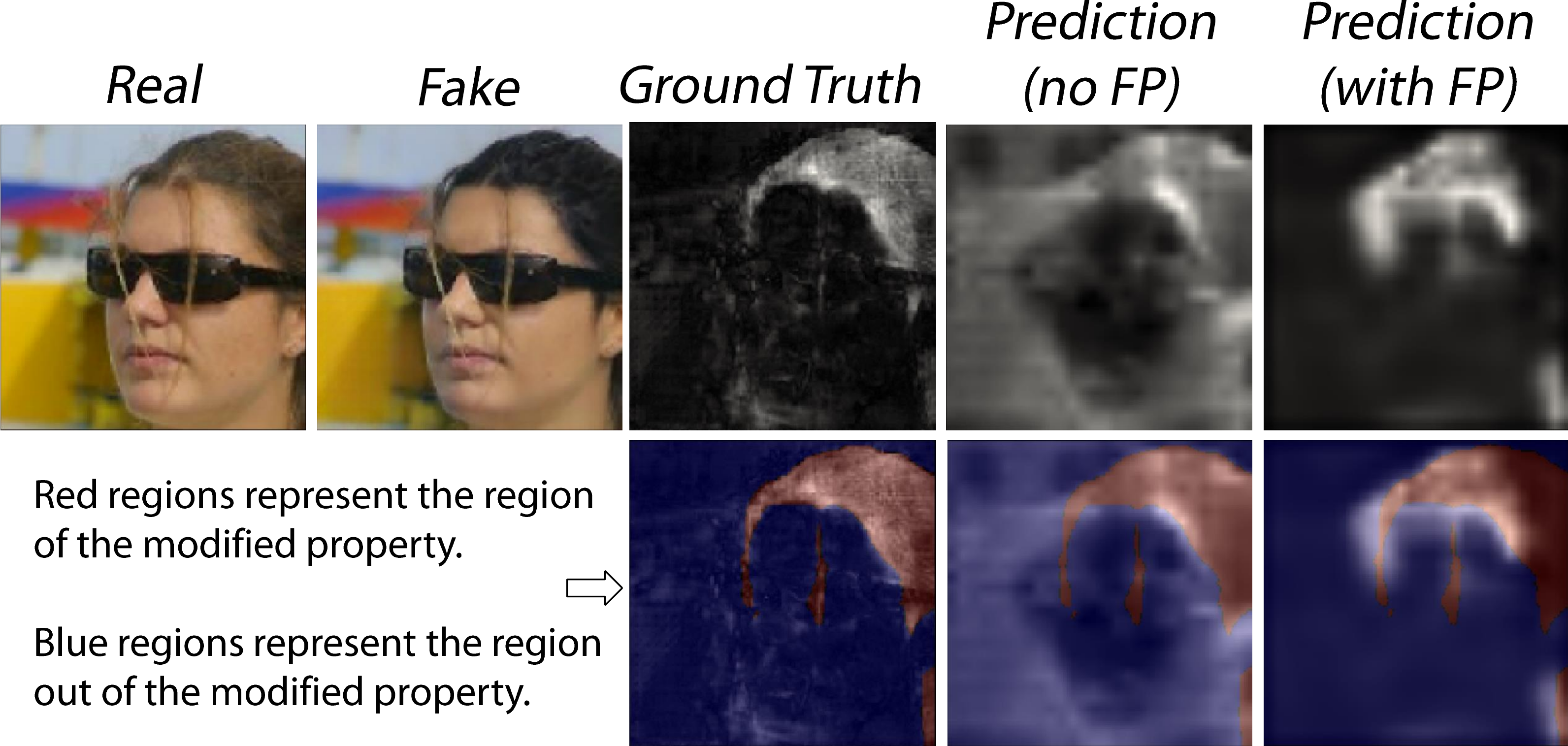}
	\caption{Result of the models (model with face parsing and model without face parsing) testing on STGAN (Black hair). The models are trained on STGAN (Blond hair). The first row in turn shows the real image, fake image, ground truth, prediction without face parsing, prediction with face parsing. The second row shows the regions where we used to calculate similarity metrics.}
	\label{fig:limited_image_method_performance}
\end{figure}

As introduced in Fig.~\ref{fig:framework}, through adding face attention information, the model can learn fake texture better. In the testing procedure, we use a white map that does not provide any location information as the face attention map. The result shows that the model does learn fake texture better. As an example, we train two models on STGAN (Blond hair). One has face attention information while the other does not. In Fig.~\ref{fig:limited_image_method_performance}, we demonstrate the localization performance of the model testing on STGAN (Black hair). \textbf{FP} means face parsing. The model with face parsing achieves a better result. It can capture the main fake textures while the model without face parsing is in a mess. 

\begin{figure}[tbp]
	\centering
	\includegraphics[width=\columnwidth]{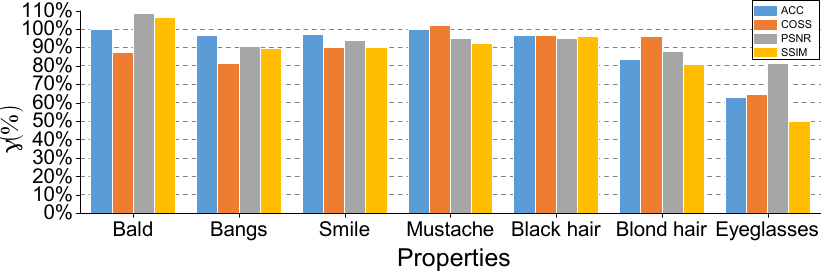}
	\caption{Comparison with no disorganization. $\gamma$ means the percentage of the metric values compared with no disorganization.}
	\label{fig:robustness_compare}
\end{figure}

Because this is a cross-attribute test, so we use in-region similarity to compare the results with ground truth. In-region means the region of the modified property. In Fig.~\ref{fig:limited_image_method_performance}, it represents the red region of the image in the second row.

In Table~\ref{tab:compare_face_parsing}, we show the testing performance on STGAN (Black hair), STGAN (Brown hair), and STGAN (Smile) of the model trained by STGAN (Blond hair). The similarity metrics between the model with face parsing or the model without face parsing are similar. However, the detection accuracy of the model with face parsing is much better than that without face parsing. In Fig.~\ref{fig:limited_image_method_performance}, we can also find that model with parsing can catch fake regions more accurately. To the model without face parsing, if we change facial property \emph{Hair} with different colors, the cross-attribute universality of it will reduce. To the model with face parsing, it learns fake texture better with the introduction of the attention mechanism, which leads to a better cross-attribute universality. It even gets a good detection accuracy on facial property \emph{Smile}, which is very different from facial property \emph{Hair}.

\subsubsection{Robustness of the model against degradations}
It is very necessary to test the robustness of the localization model. In the real world, images may be degraded by various operations such as compression, low-resolution, \etc. Moreover, we also need to ensure that the model can locate fake textures in any place. This means that the model is location-independent.

\begin{figure}[tbp]
	\centering 
	\includegraphics[width=\columnwidth]{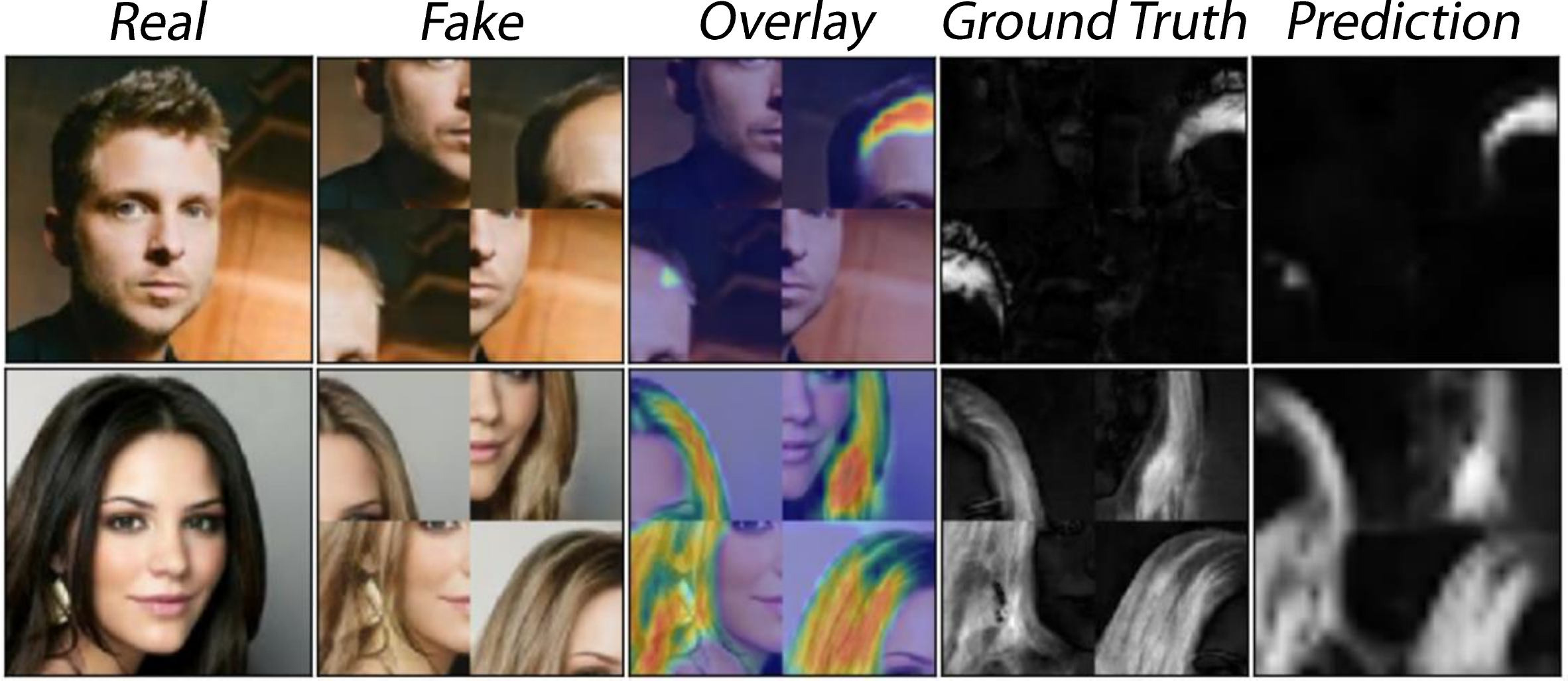}
	\caption{\textbf{Disorganization test result.} These are the property \emph{Bald} and \emph{Blond hair} of STGAN. We clip the image into four pieces and mess up their order. This operation is for verifying whether our model is location-independent. The result shows that our model is robust to this test.}
	\label{fig:robustness_of_model}
\end{figure}

To test the robustness of our model, we use two different test scenarios. In the first scenario, we crop each fake image uniformly into four pieces and splice them randomly, which is called \emph{disorganization test}. The purpose of this test is to verify whether the model is strongly correlated to the location, namely, it just remembers the location instead of recognizing the fake texture. Fig.~\ref{fig:robustness_compare} shows the percentage of the metric values compared to that before the disorganization test. The model performance is just a little worse than the formal situation. Fig.~\ref{fig:robustness_of_model} shows the prediction result of the \emph{disorganization test}.

\begin{figure}[tbp]
\centering
\subfigure[JPEG Compression]{
\includegraphics[width=0.44\columnwidth]{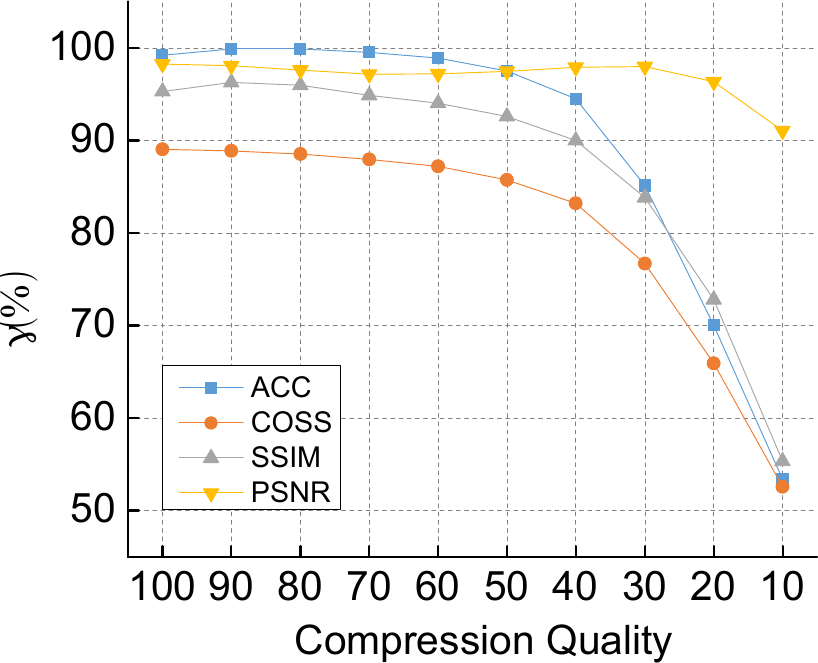}
}
\quad
\subfigure[Blur]{
\includegraphics[width=0.44\columnwidth]{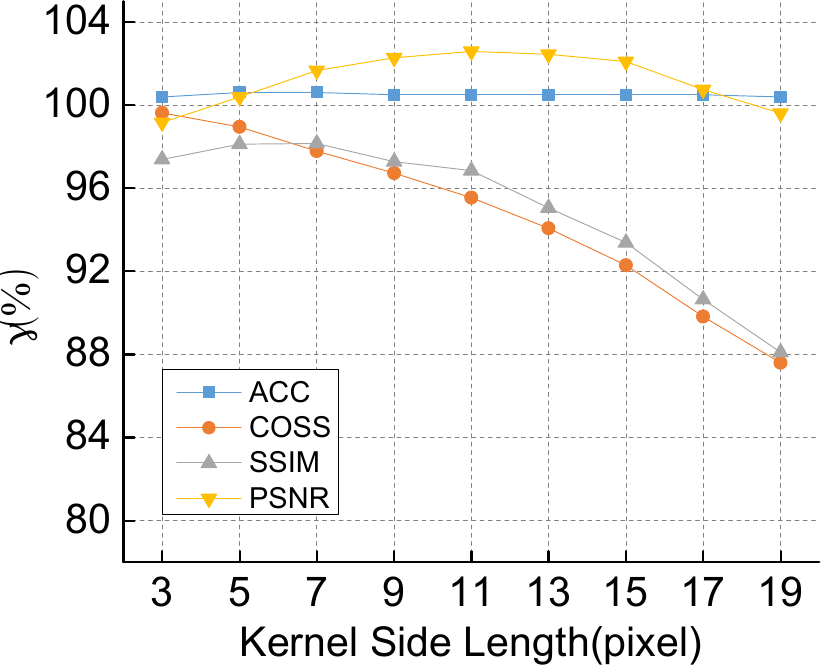}
}
\quad
\subfigure[Noise]{
\includegraphics[width=0.44\columnwidth]{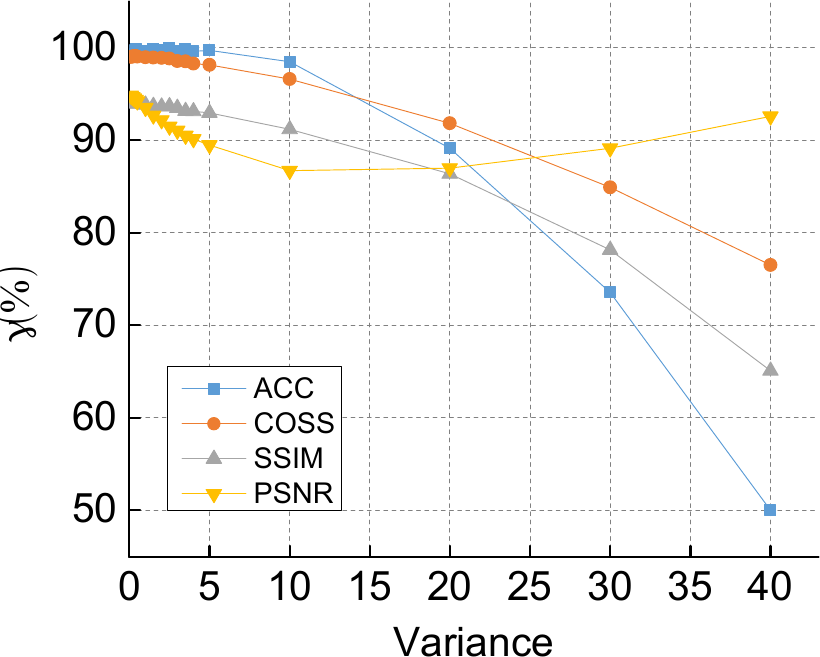}
}
\quad
\subfigure[Low-resolution]{
\includegraphics[width=0.44\columnwidth]{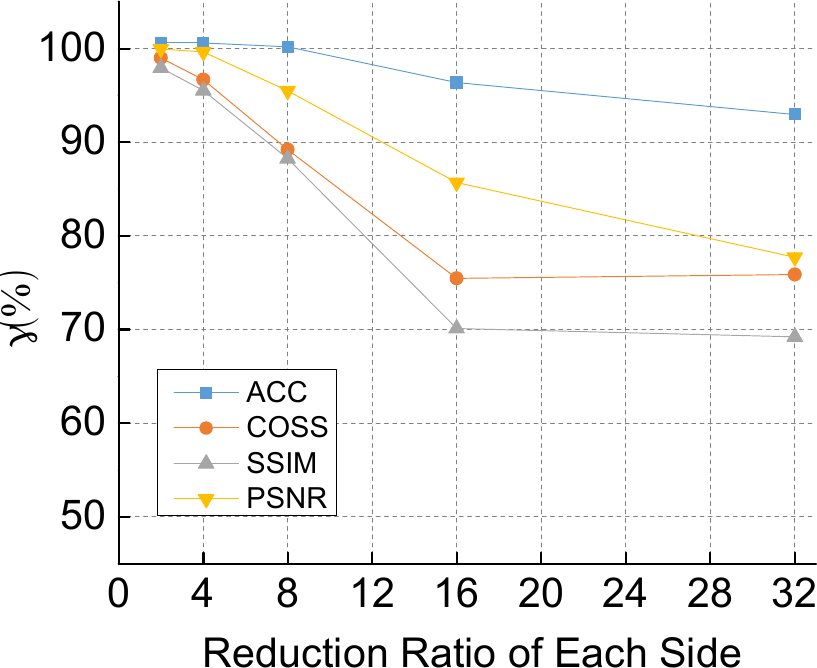}
}
\caption{The anti-degradation capability of the model. $\gamma$ means the percentage of the metric values compared with no degradation.}
\label{fig:degradation}
\end{figure}

\begin{figure*}
	\centering 
	\includegraphics[width=\linewidth]{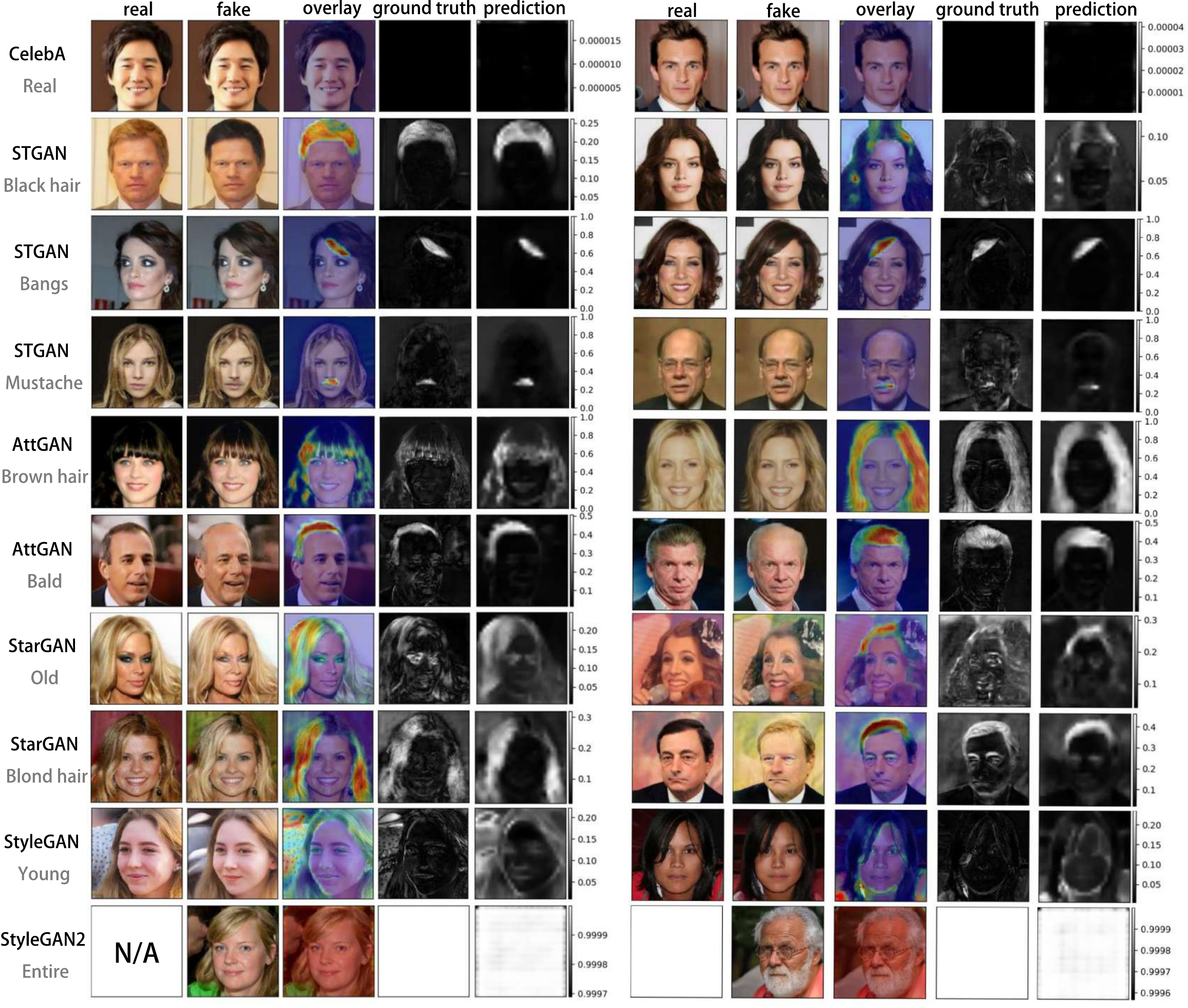}
	\caption{Fake region localization results. These are the results of different GANs and properties. In the left comments, CelebA \cite{liu2018large} is a real image database and others are GAN-based face generation methods. The gray text represents the facial property. 
	\textbf{Fake} image is produced by manipulating corresponding \textbf{real} image through GAN-based face generation method. \textbf{Ground truth} is calculated by \textbf{fake} image and \textbf{real} image. \textbf{Fake} image and \textbf{ground truth} are the input and expected output of our method. \textbf{Overlay} is combined of \textbf{prediction} and \textbf{fake} image. The \textbf{prediction} has colorbar which shows the value range of pixels. For the first row which uses real image of CelebA as input, for unity of the figure, we also regard it as \textbf{fake} image.}
	\label{fig:show-big-image}
\end{figure*}
In the second scenario, we also apply four different real-world facial image degradations (\textbf{JPEG Compression, Blur, Noise, and Low-resolution}) on 1,000 fake images. The gray-scale fakeness ground truth map is processed by the real image and the degraded fake image. In Fig.~\ref{fig:degradation}, the vertical axis of all the four subfigures represents the percentage of the metric values compared to that before degradation. \textbf{JPEG Compression} means that converting an image from PNG format to JPEG format. The horizontal axis of the image represents the compression quality during conversion. \textbf{Blur} and \textbf{Noise} mean that applying Gaussian blur and Gaussian noise to fake images respectively. The horizontal axis respectively represents the filter size of the Gaussian blur and the variance of the Gaussian noise. \textbf{Low-resolution} means that resizing the fake image to a low resolution, then restoring it to the original resolution with linear interpolation. The quality of the fake image reduces in the resizing procedure. The horizontal axis represents the reduction ratio of each side of the image.

We can find that all the metric values gradually reduce in \textbf{Low-resolution}. In the other three degradations, \textbf{PSNR} achieves only minor changes, while \textbf{ACC}, \textbf{COSS}, and \textbf{SSIM} decrease when the interference is extremely high. Overall, our methodology performs well against various degradations.

\section{Conclusion}
In this paper, we utilize the imperfection of the upsampling procedure in all the GAN-based partial face manipulation methods and entire face synthesis methods. This imperfection can be used for fake detection and fake localization. Thus we propose a universal pipeline to solve the fake localization problem. Through using a gray-scale fakeness map, we achieve the SOTA localization accuracy. As an improvement of the cross-attribute universality of the model, the attention mechanism is inserted into the encoder-decoder architecture by using face parsing information. We also propose partial data augmentation and single sample clustering to improve the cross-method universality of our model. The model is robust to real-world degradations such as blur, compression, \etc.

Beyond DeepFake detection and localization, we conjecture that the \emph{FakeLocator} is capable of detecting and localizing non-additive noise adversarial attacks such as \cite{arxiv19_amora,guo2020watch} where the attacked images do not exhibit visual noise patterns and are usually much harder to detect accurately. Recently, some works also have tried to improve the fake images by reconstruction methods \cite{huang2020deepnotch,huang2020fakepolisher}, which propose a new challenge to our method. In future work, we think visualizing the fake texture in each image and classify them according to different GANs and upsampling methods is a good idea.


%



\ifCLASSOPTIONcaptionsoff
  \newpage
\fi



%

\bibliographystyle{IEEEtran}
\bibliography{ref}

%




\end{document}